\documentclass[11pt, hidelinks]{article} % For LaTeX2e

\usepackage{hyperref}
\usepackage{url}
\usepackage{fullpage}
\usepackage{xcolor}
\usepackage{enumitem, comment, xifthen}
\usepackage{graphicx}
\graphicspath{{figs/}}
\usepackage{subfig}
\usepackage{amsmath,amssymb,amsthm, amsfonts}
\usepackage[T1]{fontenc}

\theoremstyle{plain}

\newtheorem{theo}{Theorem}

\newtheorem{lem}{Lemma}
\newtheorem{prop}{Proposition}
\newtheorem{cor}{Corollary}

\theoremstyle{definition} 

\newtheorem{nota}{Notation}
\newtheorem{de}{Definition}
\newtheorem{exa}{Example}
\newtheorem{as}{Assumption}
\newtheorem{alg}{Algorithm}

\newcommand{\btheo}{\begin{theo}}
\newcommand{\bde}{\begin{de}}
\newcommand{\ble}{\begin{lem}}
\newcommand{\bpr}{\begin{prop}}
\newcommand{\bno}{\begin{nota}}
\newcommand{\bex}{\begin{exa}}
\newcommand{\bcor}{\begin{cor}}
\newcommand{\spro}{\begin{proof}}
\newcommand{\bas}{\begin{as}}
\newcommand{\balg}{\begin{alg}}

\newcommand{\etheo}{\end{theo}}
\newcommand{\ede}{\end{de}}
\newcommand{\ele}{\end{lem}}
\newcommand{\epr}{\end{prop}}
\newcommand{\eno}{\end{nota}}
\newcommand{\eex}{\end{exa}}
\newcommand{\ecor}{\end{cor}}
\newcommand{\fpro}{\end{proof}}
\newcommand{\eas}{\end{as}}
\newcommand{\ealg}{\end{alg}}

\theoremstyle{plain}

\newtheorem{theos}{Theorem}
\newtheorem{props}{Proposition}
\newtheorem{lems}{Lemma}
\newtheorem{cors}{Corollary}

\theoremstyle{definition}
\newtheorem{exas}{Example}
\newtheorem{algs}{Algorithm}
\newtheorem{asss}{Assumption}
\newtheorem{defns}{Definition}

\newcommand{\btheos}{\begin{theos}}
\newcommand{\etheos}{\end{theos}}
\newcommand{\bprops}{\begin{props}}
\newcommand{\eprops}{\end{props}}
\newcommand{\bdes}{\begin{defns}}
\newcommand{\edes}{\end{defns}}
\newcommand{\blems}{\begin{lems}}
\newcommand{\elems}{\end{lems}}
\newcommand{\bcors}{\begin{cors}}
\newcommand{\ecors}{\end{cors}}
\newcommand{\bexs}{\begin{exas}}
\newcommand{\eexs}{\end{exas}}
\newcommand{\balgs}{\begin{algs}}
\newcommand{\ealgs}{\end{algs}}
\newcommand{\bass}{\begin{asss}}
\newcommand{\eass}{\end{asss}}

\newcommand{\argmin}{\mathop{\rm arg\,min}}

\newcommand{\T}{\mathsf{T}}%\mathrm{T}} % inner product.
\makeatletter
\newcommand*{\coloneq}{\mathrel{\rlap{%
                     \raisebox{0.3ex}{$\m@th\cdot$}}%
                     \raisebox{-0.3ex}{$\m@th\cdot$}}%
  =}
\makeatother
\makeatletter
\newcommand*{\eqcolon}{=\mathrel{\rlap{%
                     \raisebox{0.3ex}{$\m@th\cdot$}}%
                     \raisebox{-0.3ex}{$\m@th\cdot$}}%
  }
\makeatother
\newcommand{\ceil}[1]{\left\lceil #1 \right\rceil}

\newenvironment{enummath}
 {\begin{enumerate}[font=\upshape,label=(\alph*)]}
 {\end{enumerate}}

\newcommand{\R}{\mathbf R} % reals
\newcommand{\e}{\mathrm{e}}

\newcommand{\iid}{i.i.d.}
\DeclareMathOperator{\Lip}{Lip}

\DeclareMathOperator{\diag}{\bf diag}

\DeclareMathOperator{\prox}{\bf prox}

\newcommand{\cT}{\mathcal{T}}

\newcommand{\1}{\mathbf 1} % ones
\let\ones\1
\newcommand{\half}{\frac12}

\newcommand{\simiid}{\overset{\text{i.i.d.}}{\sim}}
\newcommand{\simind}{\overset{\textrm{ind.}}{\sim}}

\DeclareMathOperator{\PP}{\bf P}
\renewcommand{\P}{\PP}

\let\hat\widehat

\renewcommand{\leq}{\leqslant}
\renewcommand{\geq}{\geqslant}

\newcommand{\ie}{\textit{i}.\textit{e}., }
\newcommand{\eg}{\textit{e}.\textit{g}., }

\newcounter{algorithmctr}
\renewcommand{\thealgorithmctr}{\arabic{algorithmctr}}
\newenvironment{algdesc}%
   {\refstepcounter{algorithmctr}\begin{list}{}{%
       \setlength{\rightmargin}{0\linewidth}%
       \setlength{\leftmargin}{0\linewidth}}%
       \rmfamily\small
       \item[]{\setlength{\parskip}{0ex}\hrulefill\par%
        \nopagebreak{\bfseries\textsf{Algorithm \thealgorithmctr~}}}}%
   {{\setlength{\parskip}{-1ex}\nopagebreak\par\hrulefill} \end{list}}

\makeatletter
\long\def\@makecaption#1#2{
        \vskip 0.8ex
        \setbox\@tempboxa\hbox{\small {\bf #1.} #2}
        \parindent 1.5em 
        \dimen0=\hsize
        \advance\dimen0 by -3em
        \ifdim \wd\@tempboxa >\dimen0
                \hbox to \hsize{
                        \parindent 0em
                        \hfil 
                        \parbox{\dimen0}{\def\baselinestretch{0.96}\small
                                {\bf #1.} #2
                                } 
                        \hfil}
        \else \hbox to \hsize{\hfil \box\@tempboxa \hfil}
        \fi
        }
\makeatother

\newcommand{\xstar}{\ensuremath{x^\star}}
\newcommand{\gradop}{\nabla F} %\ensuremath{\cT_{\rm grad}}}
\newcommand{\eqop}{\mathcal{N}_\eqset} %\ensuremath{\cT_{\rm eq}}}
\newcommand{\numepochs}{e}
\newcommand{\numdevices}{m}
\newcommand{\numdevice}{\numdevices}

\newcommand{\gradj}{G_j}
\newcommand{\xls}{\xstar_{\rm ls}}
\newcommand{\xfed}{\xstar_{\rm Fed}}
\newcommand{\xfedprox}{\xstar_{\texttt{FedProx}}}
\newcommand{\xfedgrad}{\xstar_{\texttt{FedGD}}}
\newcommand{\step}{\ensuremath{s}}

\newcommand{\gstep}{\ensuremath{\alpha}}
\newcommand{\xtrue}{\ensuremath{x_{0}}} 
\newcommand{\noisevec}{\ensuremath{v}}
\newcommand{\Normal}[2]{\ensuremath{\mathsf{N}\left(#1, #2 \right)}}

\newcommand{\Fstar}{\ensuremath{F^\star}}
\let\eps\varepsilon
\let\epsilon\varepsilon

\newcommand{\tildeO}[1]{\widetilde{O}\left(#1\right)}
\newcommand{\zinit}{x} % strangely, I think we actually want to interpret the initialization as an x-init rather than z-init. 
\newcommand{\Lmax}{\ensuremath{L^\ast}}
\newcommand{\Lsum}{\overline{L}}
\newcommand{\ellmin}{\ensuremath{\ell_\ast}}
\newcommand{\proxjplain}{ {\mathtt{prox\_update}}_j}
\newcommand{\proxj}[1]{\ensuremath{\proxjplain(#1)}}
\newcommand{\zfp}{z^\star}
\DeclareMathOperator{\refl}{\bf refl}
\newcommand{\aprox}{\widetilde{\prox}}
\newcommand{\aproxj}[2]{\mathtt{prox\_update}_{#1}(#2)}
\newcommand{\arefl}{\widetilde{\refl}}
\newcommand{\eqset}{E}
\newcommand{\Orthog}{\ensuremath{\mathrm{O}}}
\newcommand{\zstar}{z^\star}
\let\phi\varphi
\newcommand{\Unif}{\ensuremath{\mathsf{Unif}}}
\newcommand{\numepoch}{\numepochs}
\newcommand{\defn}{\ensuremath{\coloneq}}

\newcommand{\real}{\ensuremath{\mathbf{R}}}

\newcommand{\upstairs}[1]{\textsuperscript{#1}}
\newcommand{\affilone}{\dag}
\newcommand{\affiltwo}{\ddag}
\newcommand{\affilthree}{\ddag\ddag}

\newcommand{\fedsplit}{\texttt{FedSplit}}
\newcommand{\fedsplitspace}{\fedsplit$\:$}

\newcommand{\mns}{\mkern-1mu}  % My negative space
\newcommand{\opnorm}[1]{|\mns|\mns| #1 |\mns|\mns|_{\mbox{\tiny{op}}}}
\newcommand{\ustar}{\ensuremath{u^*}}

\newcommand{\That}{\ensuremath{ \hat \cT}}

\newcommand{\usedim}{\ensuremath{d}}

\newcommand{\widgraph}[2]{\includegraphics[keepaspectratio,width=#1]{#2}}

\newcommand{\xhat}{\ensuremath{\widehat{x}}}

\newcommand{\Tfedgrad}{\ensuremath{T_{\rm FedGrad}}}
\newcommand{\Tfedsplit}{\ensuremath{T_{\rm FedSplit}}}

\newcommand{\numobs}{n}

\begin{document}

\begin{center}

  {\bf{\Large \fedsplit: An algorithmic framework for fast federated optimization}} \\

  \vspace*{.2in}
  
  \begin{tabular}{cc}
    Reese Pathak\upstairs{\affilone}
    and
    Martin J.\ Wainwright\upstairs{\affilone, \affiltwo, \affilthree} \\
    \upstairs{\affilone} Department of Electrical Engineering and Computer Science, UC Berkeley \\
    \upstairs{\affiltwo} Department of Statistics, UC Berkeley \\
    \upstairs{\affilthree} Voleon Group, Berkeley \\
   \texttt{\string{pathakr,wainwrig\string}@berkeley.edu}
  \end{tabular}
  
  \vspace*{.2in}

  \begin{abstract}
    Motivated by federated learning, we consider the hub-and-spoke
    model of distributed optimization in which a central authority
    coordinates the computation of a solution among many agents while
    limiting communication.  We first study some past procedures for
    federated optimization, and show that their fixed points need not
    correspond to stationary points of the original optimization
    problem, even in simple convex settings with deterministic
    updates.  In order to remedy these issues, we introduce \fedsplit,
    a class of algorithms based on operator splitting procedures for
    solving distributed convex minimization with additive structure.
    We prove that these procedures have the correct fixed points,
    corresponding to optima of the original optimization problem, and
    we characterize their convergence rates under different settings.
    Our theory shows that these methods are provably robust to inexact
    computation of intermediate local quantities.  We complement our
    theory with some simple experiments that demonstrate the benefits
    of our methods in practice.
  \end{abstract}
\end{center}

%%%%%%%%%%%%%%%%%%%%%%%%%%%%%%%%%%%%%%

\section{Introduction}

Federated learning is a rapidly evolving application of distributed
optimization for estimation and learning problems in large-scale
networks of remote clients~\cite{KaiEtAl19}. These systems present new
challenges, as they are characterized by heterogeneity in
computational resources and data across the network, unreliable
communication, massive scale, and privacy
constraints~\cite{LiEtAl19}. A typical application is for
developers of cell phones and cellular applications
to model the usage of software
and devices across millions or even billions of users.

Distributed optimization has a rich history and extensive literature
(e.g., see the sources~\cite{BerTsi, BoydEtAl10, DekEtAl12,
  ZhaDucWai13, LiEtAl14, RicTak16} and references therein), and
federated learning has led to a flurry of interest in the area.  A
number of different procedures have been proposed for federated
learning and related problems, using methods based on stochastic
gradient methods or proximal procedures. Notably, McMahan et
al.~\cite{McMaEtAl16} introduced the \texttt{FedSGD} and
\texttt{FedAvg} algorithms, which both adapt the classical stochastic
gradient method to the federated setting, considering the possibility
that clients may fail and may only be subsampled on each round of
computation. Another recent proposal has been to use regularized local
problems to mitigate possible issues that arise with device
heterogeneity and failures~\cite{LiEtAl18}.  These authors propose the
\texttt{FedProx} procedure, an algorithm that applied averaged
proximal updates to solve federated minimization problems.

Currently, the convergence theory and correctness of these
methods is currently lacking, and practitioners have documented
failures of convergence in certain settings (\eg see
Figure 3 and related discussion in the work~\cite{McMaEtAl16}).
Our first contribution
in this paper is to analyze the deterministic analogues of these
procedures, in which the gradient or proximal updates are performed
using the full data at each client; such updates can be viewed as the
idealized limit of a minibatch update based on the entire local
dataset.  Even in this especially favorable setting, we show that most
versions of these algorithms fail to preserve the fixed points of the
original optimization problem: that is, even if they converge, the
resulting fixed points need not be stationary.  Since the stochastic
updates implemented in current practice are randomized versions of the
underlying deterministic procedures, they also fail to preserve the
correct fixed points in general.

In order to address this issue, we show how operator splitting
techniques~\cite{BoydEtAl10, RyuBoyd16, Com18, BauCom17} can be
exploited to permit the development of provably correct and
convergence procedures for solving federated problems.  Concretely, we
propose a new family of federated optimization algorithms, that we
refer to as \fedsplit.  These procedures us to solve distributed
convex minimization problems of the form
\begin{align}
  \label{prob:finite-sum-problem}
\text{minimize} \quad F(x) \coloneq \sum_{j=1}^\numdevices f_j(x),
\end{align}
where $f_j \colon \R^d \to \R$ are cost functions that each client
assigns to the optimization variable $x \in \R^d$. In machine learning
applications, the vector $x \in \real^d$ is a parameter of a
statistical model. In this paper, we focus on the case when $f_j$ are
finite convex functions, with Lipschitz continuous gradient.  While
such problems are pervasive in data fitting applications, this
necessarily precludes the immediate application of our methods and
guarantees to constrained, nonsmooth, and nonconvex problems. We leave
the analysis and development of such methods to future work.

As previously mentioned, distributed optimization is not a new
discipline, with work dating back to the 1970s and
1980s~\cite{BerTsi}.  Over the past decade, there has been a
resurgence of research on distributed optimization, specifically for
learning problems~\cite{BoydEtAl10, DekEtAl12, ZhaDucWai13, LiEtAl14,
  RicTak16}.  This line of work builds upon even earlier study of
distributed first- and second-order methods designed for optimization
in the ``data center'' setting~\cite{BerTsi}.  In these
applications, the devices that carry out the computation are high
performance computing clusters with computational resources that are
well-known. This is in contrast to the federated setting, where the
clients that carry out computation are cell phones or other
computationally-constrained mobile devices for which carrying out
expensive, exact computations of intermediate quantities may be
unrealistic. Therefore, it is important to have methods that permit
approximate computation, with varying levels of accuracy throughout
the network of participating agents~\cite{BonEtAl19}.

Our development makes use of a long line of work that adapts the
theory of operator-splitting methods to distributed
optimization~\cite{BoydEtAl10, RyuBoyd16, Com18, BauCom17}.  In
particular, the \fedsplitspace procedure developed in this paper is
based upon an application of the Peaceman-Rachford
splitting~\cite{PeaceRach55} to the distributed convex
problem~\eqref{prob:finite-sum-problem}.  This method and its variants
have been studied extensively in the general setting of root-finding
for the sum of two maximally monotone operators.  Recent works have
studied such splitting schemes for convex minimization under strong
convexity and smoothness assumptions~\cite{GisBoy17, Gis17, MouVan19}.
In this paper, we adapt this general theory to the specific setting of
federated learning, and extend it to apply beyond strongly convex
losses. Furthermore, we extend this previous work to the setting when
specific intermediate quantities---likely to dominate the
computational cost of on-device training---are inexact and cheaply
computed.

The remainder of this paper is organized as follows. We begin with a
discussion of two previously proposed methods for solving federated
optimization problems in Section~\ref{sec:past-algs}.  We show that
these methods cannot have a generally applicable convergence theory as
these methods have fixed points that are not solutions to the
federated optimization problems they are designed to solve. In
Section~\ref{sec:our-alg}, we present the \fedsplitspace procedure,
and demonstrate that, unlike some other methods in use, it has fixed
points that do correspond to optimal solutions of the
original federated optimization problem.  After presenting convergence
results, we present numerical experiments in
Section~\ref{sec:numerics}.  These experiments confirm our theoretical
predictions and also demonstrate that our methods enjoy favorable
scaling in the problem conditioning.  Section~\ref{SecProofs} is
devoted to the proofs of our results. We conclude in
Section~\ref{sec:discussion} with future directions suggested by the
development in this paper.

%%%%%%%%%%%%%%%%%%%%%%%%%%%%%%%%%%%%%%%%%%%%%%%%%%%%%%%%%%%%%%%%%%%%%%%%%%%%%%%%%%%%%%%%%%%%%%%%%%%%%%%

\subsection{Notation} 

For the reader's convenience, we collect here our notational conventions.

\paragraph{Set and vector arithmetic:}

Given vectors $x, y \in \real^d$, we use $x^\T y = \sum_{j=1}^d x_j
y_j$ to denote their Euclidean inner product, and $\|x \| =
\sqrt{x^\T x}$ to denote the Euclidean norm.  Given two non-empty
subsets $A, B \subset \R^d$, their Minkowski sum is given by $A + B =
\{ x + y \mid x\in A, y \in B \}$.  We also set $x + B = \{x\} + B$
for any point $x \in \R^d$.

For an integer $\numdevices \geq 1$, we use the shorthand
$[\numdevices] \defn \{1, \ldots, \numdevices \}$.  Given a
block-partitioned vector $z = (z_1, \dots, z_\numdevices) \in
(\real^d)^\numdevices$ with $z_j \in \R^d$ for $j \in [\numdevices]$,
we define the block averaged vector \mbox{$\overline{z} \coloneq
  \frac{1}{\numdevices} \sum_{j=1}^\numdevices z_j$.}  Very
occasionally, we also slightly abuse notation by defining arithmetic
between vectors of dimension with a common factor.  For example, if $x
\in \R^d$ and $(y_1, \dots, y_\numdevices) = y \in
(\R^{d})^{\numdevices}$, then
\begin{align*}
x + y \coloneq (x + y_1, \dots, x + y_\numdevices).
\end{align*}

\paragraph{Regularity conditions:} A differentiable function $f\colon \R^d \to \R$ is said to be
$\ell$-strongly convex if
\begin{align*}
f(y) & \geq f(x) + \nabla f(x)^\T(y - x) + \frac{\ell}{2}\|y-x\|^2,
\quad \text{for all} \quad x, y \in \R^d.
\end{align*}
It is simply convex if this condition holds with $\ell = 0$.
Similarly, a differentiable function $f \colon \R^d \to \R$ is
$L$-smooth if its gradient $\nabla f$ is $L$-Lipschitz continuous,
\begin{align*}
\|\nabla f(x) - \nabla f(y) \| \leq L \|x - y\|, \quad \text{for
  all}~x,y \in \R^d.
\end{align*}

\paragraph{Operator notation:}
Given an operator $\cT \colon \R^d \to \R^d$ and a positive integer
$k$, we use $\cT^k$ to denote the composition of $\cT$ with itself $k$
times---that is, $\cT^k$ is a new operator that acts on a given $x \in
\real^d$ as $\cT^{k} x \coloneq \underbrace{\cT \circ \cT \circ \cdots
  \circ \cT}_{k~\text{times}} x$.
An operator $\cT \colon \R^d \to \R^d$ is said to be 
\emph{monotone} if
\begin{align}
  \label{EqnMonotone}
\left(\cT y - \cT x \right)^{\T\!}(y - x) \geq 0 \qquad \mbox{for all
  $x, y \in \real^d$.}
\end{align}

%%%%%%%%%%%%%%%%%%%%%%%%%%%%%%%%%%%%%%%%%%%%%%%%%%%%%%%%%%%%%%%%%%%%%%%%%%%%%%%%%%%%%%%%%%%%%%%%

\section{Existing algorithms and their fixed points}
\label{sec:past-algs}

Prior to proposing our own algorithms, let us discuss the fixed points
of the deterministic analogues of some methods recently proposed for
federated optimization problems~\eqref{prob:finite-sum-problem}.  We
focus our discussion on two recently proposed procedures---namely,
\texttt{FedSGD}~\cite{McMaEtAl16} and
\texttt{FedProx}~\cite{LiEtAl18}.

In understanding these and other algorithms, it is convenient to
introduce the consensus reformulation of the distributed
problem~\eqref{prob:finite-sum-problem}, which takes the form
\begin{align}
  \label{prob:consensus-reformulation}
    \begin{array}{ll}
   \text{minimize} & F(x) \coloneq \sum_{j=1}^\numdevices f_j(x_j)
   \\ \text{subject to} & x_1 = x_2 = \cdots = x_\numdevices.
    \end{array}
\end{align}
Although this consensus formulation involves more variables, it is
more amenable to the analysis of distributed procedures~\cite{BoydEtAl10}.

\subsection{Federated gradient algorithms} 
\label{sec:fedgrad}

The recently proposed \texttt{FedSGD} method~\cite{McMaEtAl16} is
based on a multi-step projected stochastic gradient method for solving
the consensus problem.  Given the iterates $\{x_j^{(t)}, j = 1,
\ldots, \numdevices\}$ at iteration $t$, the method is based on taking
some number $\numepochs \geq 1$ of stochastic gradient steps with
respect to each loss $f_j$, and then passing to the coordinating agent
to compute an average, which yields the next iterate in the sequence.
When a single stochastic gradient step ($\numepoch = 1$) is taken
between the averaging steps, this method can be seen as a variant of
projected stochastic gradient descent for the consensus
problem~\eqref{prob:consensus-reformulation} and by classical theory
of convex optimization enjoys convergence guarantees
\cite{JulSchmBach12}.  On the other hand, when the number of epochs
$\numepochs$ is strictly larger than 1, it is unclear \emph{a priori}
if the method should retain the same guarantees.

As we discuss here, even without the additional inaccuracies
introduced by using stochastic approximations to the local gradients,
\texttt{FedSGD} with $\numepochs > 1$ will not converge to minima in
general.  More precisely, let us consider the deterministic version of
this method (which can be thought of the ideal case that would be
obtained when the mini-batches at each device are taken to infinity).
Given a stepsize $\step > 0$, define the \emph{gradient mappings}
\begin{align}
  \gradj(x) \coloneq x - \step \nabla f_j(x) \qquad \mbox{ for $j = 1,
    \dots, \numdevices$.}
\end{align}
For a given integer $\numepochs \geq 1$, we define the
$\numepochs$-fold composition
\begin{align}
\gradj^\numepochs(x) & \defn \underbrace{\big(\gradj \circ \gradj
  \circ \ldots \circ \gradj \big)}_{\mbox{$\numepoch$-times}} (x)
\qquad \mbox{for all $x \in \real^d$,}
\end{align}
corresponding to taking $\numepochs$ gradient steps from a given point
$x$.  We also define $\gradj^0$ to be the identity operator---that is,
$\gradj^0(x) \defn x$ for all $x \in \real^d$.

In terms of these operators, we can define a family of
\texttt{FedGD}$(\step, \numepoch)$ algorithms, with each algorithm
parameterized by a choice of stepsize $\step > 0$ and number of
gradient rounds $\numepoch \geq 1$.  Given an initialization
$x^{(1)}$, it performs the following updates for $t = 1, 2, \ldots$:
\begin{subequations}
\label{eqn:FedGD}
\begin{align}
  x_j^{(t + 1/2)} & \coloneq \gradj^{\numepochs}(x_j^{(t)}), &
  \mbox{for $j \in [\numdevices] \defn \{1,2, \dots, \numdevices\}$,
    and}
  \label{subeqn:FedGD-recursion-one}\\
  x_j^{(t + 1)} & \coloneq \overline x^{(t+1/2)}, & \mbox{for $j \in
    [\numdevices]$,}
  \label{subeqn:FedGD-recursion-two}
\end{align}
\end{subequations}
where the reader should recall that $ \overline x^{(t+1/2)} =
\tfrac{1}{\numdevices} \sum_{j=1}^\numdevices x_j^{(t+1/2)}$ is the
block average.

\bpr
\label{prop:FedGD-fixed-points}
For any $\step > 0$ and $\numepochs
\geq 1$, the sequence $\{ x^{(t)} \}_{t=1}^\infty$ generated by the
     {\texttt{FedGD}$(\step, \numepoch)$} algorithm in
     equation~\eqref{eqn:FedGD} has the following properties:
\begin{enummath}
\item 
If $x^{(t)}$ is convergent, then the local variables $x_j^{(t)}$ share
a common limit $\xstar$ such that $x_j^{(t)} \to \xstar$ as $t \to
\infty$ for $j \in [\numdevices]$.
\item Moreover, the limit $\xstar$ satisfies the fixed point relation
\begin{align}
    \label{eqn:FedGD-fixed-point-characterization}
\sum_{i=1}^{\numepochs} \sum_{j=1}^\numdevices \nabla
f_j(G_j^{i-1}(\xstar)) = 0.
\end{align}
\end{enummath}
\epr
\noindent See Section~\ref{proof:FedGD-fixed-points} for the proof of
this claim.\\

Unpacking this claim slightly, suppose first that $\numepochs = 1$,
meaning that a single gradient update is performed at each device
between the global averaging step.  In this case, recalling that
$G_j^0$ is the identity mapping, we have $\sum_{i=1}^\numepochs \nabla
f_j (G_j^{i-1}(\xstar)) = \nabla f_j(\xstar)$, so that if $x^{(t)}$
has a limit $x$, it must satisfy the relations
\begin{align*}
x_1 = x_2 = \cdots = x_\numdevices \quad \text{and} \quad
\sum_{j=1}^\numdevices \nabla f_j(x_j) = 0.
\end{align*}
Consequently, provided that the losses $f_j$ are convex,
Proposition~\ref{prop:FedGD-fixed-points} implies that the limit of
the sequence $x^{(t)}$, when it exists, is a minimizer of the
consensus problem~\eqref{prob:consensus-reformulation}.

On the other hand, when $\numepochs > 1$, a limit of the iterate
sequence $x^{(t)}$ must satisfy the
equation~\eqref{eqn:FedGD-fixed-point-characterization}, which in
general causes the method to have limit points which are not
minimizers of the consensus problem.  For example, when $\numepochs =
2$, a fixed point $\xstar$ satisfies the condition
\begin{align*}
\sum_{j=1}^\numdevices \left \{ \nabla f_j(\xstar) + \nabla f_j \big(
\xstar - \step \nabla f_j(\xstar) \big)  \right \} & = 0.
\end{align*}
This is not equivalent to being a minimizer of the distributed problem
or its consensus reformulation, in general.

It is worth noting a very special case in which \texttt{FedGD} will
preserve the correct fixed points, even when $\numepochs > 1$.  In
particular, suppose that \emph{all} of local cost functions share a
common minimizer $\xstar$, so that $\nabla f_j(\xstar) = 0$ for 
$j \in [\numdevices]$.  Under this assumption, we have $G_j(\xstar) =
\xstar$ all $j \in [\numdevices]$, and hence by arguing inductively,
we have $G_j^i(\xstar) = \xstar$ for all $i \geq 1$.  Consequently,
the fixed point
relation~\eqref{eqn:FedGD-fixed-point-characterization} reduces to
\begin{align*}
\sum_{i=1}^{\numepochs} \sum_{j=1}^\numdevices \nabla f_j(\xstar) = 0,
\end{align*}
showing that $\xstar$ is optimal for the original federated problem.
However, the assumption that all the local cost functions $f_j$ share
a common optimum $\xstar$, either exactly or approximately, is not
realistic in practice.  In fact, if this assumption were to hold in
practice, then there would be little point in sharing data between
devices by solving the federated learning problem.

Returning to the general setting in which the fixed points need not be
preserved, let us make our observation concrete by specializing the
discussion to a simple class of distributed least squares problems.

\paragraph{Incorrectness for least squares problems:}
For $j= 1, \ldots, \numdevices$, suppose that we are given a
design matrix $A_j \in \R^{n_j \times d}$ and a response vector $b_j
\in \R^{n_j}$ associated with a linear regression problem (so that our
goal is to find a weight vector $x \in \real^d$ such that $A_j x
\approx b_j$).  The least squares regression problem defined by all
the devices takes the form
\begin{align}
  \label{prob:least-squares}
\text{minimize}\quad F(x) \defn \half \sum_{j=1}^\numdevices \|A_j x - b_j\|^2.
\end{align}
This problem is a special case of our general
problem~\eqref{prob:finite-sum-problem} with the choices
\begin{align*}
f_j(x) = \half\|A_j x - b_j\|^2, \quad j=1, \dots, \numdevices.
\end{align*}
Note that these functions $f_j$ are convex and differentiable.  For
simplicity, let us assume that the problem is nondegenerate, meaning
that the design matrices $A_j$ have full rank.  In this case, the
solution to this problem is unique, given by
\begin{align}
  \label{eqn:least-squares-solution}
  \xls = \bigg( \sum_{j=1}^\numdevices A_j^\T A_j \bigg)^{\!\!-1}
  \sum_{j=1}^\numdevices A_j^\T b_j.
\end{align}

Now suppose that we apply the \texttt{FedGD} procedure to the
least-squares problem~\eqref{prob:least-squares}.  Some
straightforward calculations yield
\begin{align*}
\nabla f_j(x) = A_j^\T (A_j x - b_j) \quad \text{and} \quad G_j(x) =
(I - \step A_j^\T A_j) x - \step A_j^\T b_j, \quad j =
1,\dots,\numdevices.
\end{align*}
Thus, in order to guarantee that $x^{(t)}$ converges, it suffices to
choose the stepsize $\step$ small enough so that $\opnorm{I - \step
  A_j^\T A_j} < 1$ for $j = 1, \ldots, \numdevices$, where
$\opnorm{\cdot}$ denotes the maximum singular value of a matrix.  In
Given the structure of the least-squares problem, the iterated
operator $G_j^k$ takes on a special form---namely:
\begin{align*}
G_j^k(x) &= (I - \step A_j^\T A_j)^k x + \step \left( \sum_{\ell =
  0}^{k-1} (I - \step A_j^\T A_j)^\ell \right) A_j^\T b_j \\ &= (I -
\step A_j^\T A_j)^k x + (A_j^\T A_j)^{-1} \left(I - (I - \step A_j^\T
A_j)^{k}\right) A_j^\T b_j.
\end{align*}
Hence, we conclude that if $x^{(t)}$ generated by the federated
gradient recursion~\eqref{subeqn:FedGD-recursion-one}
and~\eqref{subeqn:FedGD-recursion-two} converges for the least squares
problem~\eqref{prob:least-squares}, then the limit takes the form
\begin{align}
  \label{eqn:FedGD-least-squares-solution}
\xfedgrad = \Bigg(\sum_{j=1}^\numdevices A_j^\T A_j \bigg\{
\sum_{k=0}^{\numepochs - 1} (I - \step A_j^\T
A_j)^k\bigg\}\Bigg)^{\!\!-1} \Bigg(\sum_{j=1}^\numdevices
\Bigg\{\sum_{k=0}^{\numepochs - 1} (I - \step A_j^\T A_j)^{k}\bigg\}
A_j^\T b_j\Bigg).
\end{align}
Comparing this to the optimal solution $\xls$ from
equation~\eqref{eqn:least-squares-solution}), we see that as
previously mentioned when $\numepochs = 1$, that the federated
solution agrees with the optimal solution---that is, $\xfedgrad =
\xls$. However, when using a number of epochs $\numepochs > 1$ and a
number of devices $\numdevices > 1$, the fact that the coefficients in
braces in display~\eqref{eqn:FedGD-least-squares-solution} are
nontrivial implies that in general $\xfedgrad \neq \xls$.  Thus, in
this setting, federated gradient methods do not actually have the
correct fixed points, even in the idealized deterministic limit of
full mini-batches. See Section~\ref{sec:nonconvergence-leastsq} for
numerical results that confirm this observation.

%%%%%%%%%%%%%%%%%%%%%%%%%%%%%%%%%%%%%%%%%%%%%%%%%%%%%%%%%%%%%%%%%%%%%%%%%%%%%%%%%%%%%%%%%%%%%%%%%%%%%%%%%%

%%%%%%%%%%%%%%%%%%%%%%%%%%%%%%%%%%%%%%%%%%%%%%%%%%%%%%%%%%%%%%%%%%%%%%%%%%%%%%%%%%%%%%%%%%%%%%%%%%%%%%%%%%

\subsection{Federated proximal algorithms}
\label{sec:fedprox}

Another recently proposed algorithm is
\texttt{FedProx}~\cite{LiEtAl18}, which can be seen as a distributed
method loosely based on the classical proximal point
method~\cite{Roc76}.  Let us begin by recalling some classical facts
about proximal operators and the Moreau envelope; see
Rockafellar~\cite{Roc70} for more details.  For a given stepsize
$\step > 0$, the \emph{proximal operator} of a function $f \colon \R^d
\to \R$ is given by as
\begin{align}
\prox_{\step f}(z) \coloneq \argmin_{x \in \R^d} \left\{ f(x) + \frac{1}{2\step}
\|z - x\|^2 \right\}.
\end{align}
It is a regularized minimization of $f$ around $z$. The interpretation
of the parameter $\step$ as a stepsize remains appropriate in this
context: as the stepsize $\step$ grows, the penalty for moving away
from $z$ decreases, and thus, the proximal update $\prox_{\step f}(z)$
will be farther away from $z$.  When $f$ is convex, the existence of
such a (unique) minimizer follows immediately, and in this context,
the regularized problem itself carries importance:
\begin{align*}
M_{\step f}(z) \coloneq \inf_{x \in \R^d} \left\{ f(x) +
\frac{1}{2\step} \|z - x\|^2 \right\}.
\end{align*}
This function is known as the \emph{Moreau envelope} of $f$ with
parameter $\step > 0$ \cite[chap.~1.G]{RocWet09,ParEtAl14}.

With these definitions in place, we can now study the behavior of the
\texttt{FedProx} method~\cite{LiEtAl18}.  In order to bring the
relevant issues into sharp focus, let us consider a simplified
deterministic version of \texttt{FedProx}, in which we remove any
inaccuracies introduced by stochastic approximations of the gradients
(or subsampling of the devices).  For a given initialization
$x^{(1)}$, we perform the following steps for iterations $t = 1, 2,
\ldots$:
\begin{subequations}
  \begin{align}
\label{subeqn:FedProx-recursion-one}     
    x_j^{(t + 1/2)} &\coloneq \prox_{\step f_j}(x_j^{(t)}), & \quad
    \mbox{for $j \in [\numdevices]$, and} \\
 \label{subeqn:FedProx-recursion-two} 
 x_j^{(t + 1)} &\coloneq \overline x^{(t + 1/2)}, & \quad \mbox{for $j
   \in [\numdevices]$.}
  \end{align}
\end{subequations}
The following result characterizes the fixed points of this method:
\bpr
\label{prop:FedProx-fixed-points}
For any stepsize $\step > 0$, the sequence $\{x^{(t)} \}_{t=1}^\infty$
generated by the \texttt{FedProx} algorithm (see
equations~\eqref{subeqn:FedProx-recursion-one}
and~\eqref{subeqn:FedProx-recursion-two}) has the following
properties:

\begin{enummath}
\item
If $x^{(t)}$ is convergent then, the local variables $x_j^{(t)}$ share
a common limit $\xstar$ such that $x_j^{(t)} \to \xstar$ as $t \to
\infty$ for each $j \in [\numdevices]$.
\item The limit $\xstar$ satisfies the fixed point relation
  \begin{align}
\label{eqn:FedProx-fixed-point-characterization}
\sum_{j=1}^\numdevices \nabla M_{\step f_j}(\xstar) = 0.
\end{align}
\end{enummath}
\epr
\noindent See Section~\ref{proof:FedProx-fixed-points} for the proof
of this claim. \\

Hence, we see that this algorithm will typically be a zero of the sum
of the gradients of the Moreau envelopes $M_{\step f_j}$, rather than
a zero of the sum of the gradients of the functions $f_j$ themselves.
When $\numdevices > 1$, these fixed point relations are, in general,
different.

As with federated gradient schemes, one very special case in which
\texttt{FedProx} preserves the correct fixed points is when the cost
functions $f_j$ \emph{at all devices $j \in [\numdevices]$} share a
common minimizer $\xstar$. Under this assumption, the vector $\xstar$
satisfies relation~\eqref{eqn:FedProx-fixed-point-characterization}.
because the minimizers of $f_j$ and $M_{\step f_j}$ coincide, and
hence, we have $\nabla M_{\step f_j}(\xstar) = 0$ for all $j$.  Thus,
under strong regularity assumptions about the shared structure of the
device cost $f_j$, it is possible to provide theoretical guarantees
for \texttt{FedProx}.  However, as noted the assumption that the cost
functions $f_j$ all share a common optimum $\xstar$, either exactly or
approximately, is not realistic in practice.  In contrast,
the \fedsplitspace algorithm to be described in the next section
retains correct fixed points in this setting without any such
additional assumptions.

%%%%%%%%%%%%%%%%%%%%%%%%%%%%%%%%%%%%%%%%%%%%%%%%%%%%%%%%%%%%%%%%%%%%%%%%%%%%%%%%%%%%%%%%%

\paragraph{Incorrectness for least squares problems:}

In order to illustrate the fixed point
relation~\eqref{eqn:FedProx-fixed-point-characterization} from
Proposition~\ref{prop:FedProx-fixed-points} in a concrete setting, let
us return to our running example of of least squares regression. In
this setting, recall that $f_j(x) = (1/2)\|A_j x-b_j\|^2$. Thus, we
see that for any $x \in \R^d$, we have
\begin{align*}
\nabla M_{\step f_j}(x) = \frac{1}{\step} \left(x - (I + \step A_j^\T
A_j)^{-1}(x + \step A_j^\T b_j)\right), \quad j = 1, \dots, \numdevices.
\end{align*}
Thus, according to Proposition~\ref{prop:FedProx-fixed-points}, limits
$\xfed$ of the federated proximal recursion given
by~\eqref{subeqn:FedProx-recursion-one}
and~\eqref{subeqn:FedProx-recursion-two} have the form
\begin{align*}
\xfedprox = \Bigg(\sum_{j=1}^\numdevices \Big \{ I - (I + \step A_j^\T
A_j)^{-1} \Big \} \Bigg)^{\!\!-1} \bigg(\sum_{j=1}^\numdevices (A_j^\T
A_j + (1/\step)I)^{-1} A_j^\T b_j\bigg).
\end{align*}
Hence, comparing with $\xls$ as in
equation~\eqref{eqn:least-squares-solution}, in general we will have
$\xfedprox \neq \xls$.
See Section~\ref{sec:nonconvergence-leastsq} for
numerical results that confirm this observation.

%%%%%%%%%%%%%%%%%%%%%%%%%%%%%%%%%%%%%%%%%%%%%%%%%%%%%%%%%%%%%%%%%%%%%%%%%%%%%%%%%%%%%%%%%%%

\subsection{Illustrative simulation}
\label{sec:nonconvergence-leastsq}

It is instructive to perform a simple numerical experiment to see that
even in the simplest deterministic setting considered here, the
\texttt{FedProx}~\cite{LiEtAl18} and \texttt{FedSGD}~\cite{McMaEtAl16}
procedures, as specified in Sections~\ref{sec:fedprox}
and~\ref{sec:fedgrad} respectively, need not converge to the minimizer
of the original function $F$.
\begin{figure}[h!]
  \centering
  \widgraph{0.65\linewidth}{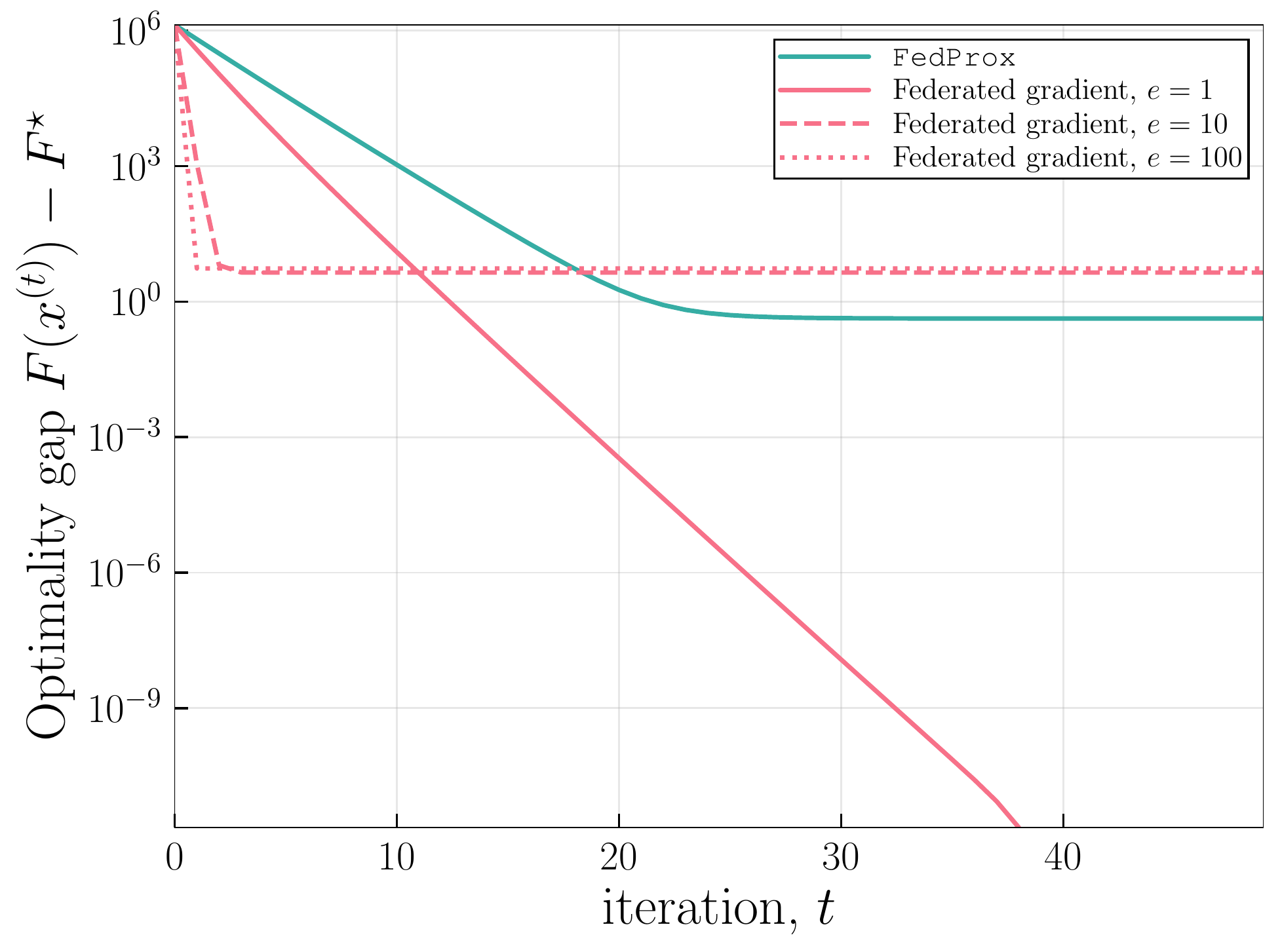}
  \caption{Plots of the optimality gap $F(x^{(t)}) - \Fstar$ versus
    the iteration number $t$ for various algorithms as applied to a
    simple least-squares problem~\eqref{prob:least-squares}.
    Here
    $\Fstar$ denotes the optimal cost value and $F(x^{(t)})$ denote
    the cost returned at round $t$ by a given algorithm.  Shown are
    curves for the \texttt{FedProx} algorithm, and the deterministic
    (infinite batch size) instantiation of \texttt{FedSGD}, with
    varying number of local epochs $\numepochs \in \{1, 10, 100\}$.
    With the exception of \texttt{FedSGD} with $\numepochs = 1$, all
    of the remaining algorithm fail to converge to an optimal
    solution, as shown by the error floors in the optimality gap.}
  \label{fig:nonconvergence}
\end{figure}

For the purposes of this illustration, we simulate an instance of our
running least squares example.  Suppose that for each device $j \in
[\numdevices]$, the response vector $b_j \in \R^{n_j}$ is related to
the design matrix $A_j$ via the standard linear model
\begin{align*}
b_j = A_j \xtrue + \noisevec_j,
\end{align*}
where $\xtrue \in \R^d$ is the unknown parameter vector to be
estimated, and the noise vectors $\noisevec_j$ are independently
distributed as $\noisevec_j \simind \Normal{0}{\sigma^2 I_{n_j}}$ for
some $\sigma > 0$.  For our experiments reported here, we constructed
a random instance of such a problem with
\begin{align*}
\numdevices = 25, \quad d = 100, \quad n_j \equiv 500, \quad
\text{and} \quad \sigma^2 = 0.25.
\end{align*}
We generated the design matrices with \iid entries of the form
$(A_j)_{kl} \simiid \Normal{0}{1}$, for $k = 1, \dots, n_j$ and $l =
1, \dots, d$.  The aspect ratios of $A_j$ satisfy $n_j > d$ for all
$j$, thus by construction the matrices $A_j$ are full rank with
probability 1.

Figure~\ref{fig:nonconvergence} shows the results of applying the
(deterministic) versions of \texttt{FedProx} and \texttt{FedSGD}, with
varying numbers of local epochs for the least squares minimization
problem~\eqref{prob:least-squares}.  As expected, we see that
\texttt{FedProx} and multi-step, deterministic \texttt{FedSGD} fail to
converge to the correct fixed point for this problem. Although the
presented deterministic variant of \texttt{FedSGD} will converge when
a single local gradient step is taken between communication rounds
(\ie when $\numepochs = 1$), we see that it also does not converge to
the optimal solution as soon as $\numepochs > 1$.

%%%%%%%%%%%%%%%%%%%%%%%%%%%%%%%%%%%%%%%%%%%%%%%%%%%%%%%%%%%%%%%%%%%%%%%%%%%%%%%%%%%%%%%%%%%%%%%%%%%%%

\section{A splitting framework and convergence guarantees}
\label{sec:our-alg}

We now turn to the description of a framework that allows us to
provide a clean characterization of the fixed points of iterative
algorithms and to propose algorithms with convergence guarantees.
Throughout our development, we assume that each function $f_j \colon
\real^d \rightarrow \real$ is convex and differentiable.

%%%%%%%%%%%%%%%%%%%%%%%%%%%%%%%%%%%%%%%%%%%%%%%%%%%%%%%%%%%%%%%%%%%%%%%%%%%%%%%%%%%%%%%%%%%%%%

\subsection{An operator-theoretic view}
\label{sec:framework}

We begin by recalling the consensus
formulation~\eqref{prob:consensus-reformulation} of the problem in
terms of a block-partitioned vector $x = (x_1, \ldots, x_\numdevices)
\in (\real^d)^\numdevices$, the function $F \colon
(\real^d)^\numdevice \rightarrow \real$ given by $F(x) \defn
\sum_{j=1}^\numdevices f_j(x_j)$, and the constraint set $\eqset
\coloneq \{x \mid x_1 = x_2 = \cdots = x_\numdevices\}$ is the
feasible subspace for problem~\eqref{prob:consensus-reformulation}.
By appealing to the first-order optimality conditions for the
problem~\eqref{prob:consensus-reformulation}, it is equivalent to find
a vector $x \in (\real^d)^\numdevice$ such that $\nabla F(x)$ belongs
to the normal cone of the constraint set $\eqset$, or equivalently
such that $\nabla F(x) \in \eqset^\perp$.  Equivalently, if we define
a set-valued operator $\eqop$ as
\begin{align}
\eqop(x) & \defn
  \begin{cases}
    \eqset^\perp, & x_1 = x_2 = \cdots = x_\numdevices, \\
    \emptyset, & \text{else} 
  \end{cases}
\end{align}
then it is equivalent to find a vector $x \in (\real^d)^\numdevices$
that satisfies the inclusion condition
\begin{align}
  \label{prob:monotone-inclusion}
  0 \in \gradop(x) + \eqop(x).
\end{align}
where $\gradop(x) = (\nabla f_1(x_1), \dots, \nabla f_m(x_m))$.

When the loss functions $f_j \colon \R^d \to \R$ are convex, both
$\gradop$ and $\eqop$ are monotone operators on $(\R^d)^\numdevices$,
as defined in equation~\eqref{EqnMonotone}.  Thus, the
display~\eqref{prob:monotone-inclusion} is a \emph{monotone inclusion
  problem}.  Methods for solving monotone inclusions have a long
history of study within the applied mathematics and optimization
literatures~\cite{RyuBoyd16, Com18}.  We now use this framework to
develop and analyze algorithms for solving the federated problems of
interest.

%%%%%%%%%%%%%%%%%%%%%%%%%%%%%%%%%%%%%%%%%%%%%%%%%%%%%%%%%%%%%%%%%%%%%%%%%%%%%%%%%%%%%%%%%%%%%%%%%%%

\subsection{Splitting procedures for federated optimization}

As discussed above, the original distributed minimization problem can
be reduced to finding a vector $x \in (\R^d)^\numdevices$ that
satisfies the monotone inclusion~\eqref{prob:monotone-inclusion}.  We
now describe a method, derived from splitting the inclusion relation,
whose fixed points do correspond with global minima of the distributed
problem.  It is an instantiation of the Peaceman-Rachford splitting,
which we refer to as the \fedsplitspace algorithm in this distributed
setting.

\begin{algdesc}[\fedsplit]
\label{alg:ourmethod}
\emph{Splitting scheme for solving federated problems of the
  form~\eqref{prob:finite-sum-problem}}
\begin{tabbing}
  {\bf Given} initialization $\zinit \in \real^d$, proximal solvers
  $\proxjplain \colon \real^d \rightarrow \real^d$\\ {\bf Initialize}
  $x^{(1)} = z^{(1)}_1 = \cdots = z^{(1)}_\numdevices = \zinit$
  \\ {\bf for} $t=1, 2,\ldots$:\\ \qquad \=\ 1.\ {\bf for} $j=1,
  \dots, \numdevices$: \\ \qquad \qquad \=\ a.\ \emph{Local prox
    step:} set $z_j^{(t + 1/2)} = \proxj{2x^{(t)} - z_j^{(t)}}$
  \\[1.3ex] \qquad \qquad \=\ b.\ \emph{Local centering step:} set
  $z_j^{(t + 1)} = z_j^{(t)} + 2(z_j^{(t+1/2)} - x^{(t)})$ \\ %[1.3ex]
  \qquad \=\ \quad~{\bf end for}\\ \qquad \=\ 2. \emph{Compute global
    average:} set $x^{(t+1)} = \overline z^{(t+1)}$. \\ {\bf end for}
\end{tabbing}
\end{algdesc}

As laid out in Algorithm~\ref{alg:ourmethod}, at each time $t = 1, 2,
\ldots$, the \fedsplitspace procedure maintains and updates a
parameter vector $z_j^{(t)} \in \real^d$ for each device $j \in
[\numdevices]$.  The central server maintains a parameter vector
$x^{(t)} \in \real^d$, which collects averages of the parameter
estimates at each machine.

The local update at device $j$ is defined in terms of a proximal
solver $\proxj{\cdot}$.  In the ideal setting, this proximal solver
corresponds to an \emph{exact} evaluation of the proximal operator
$\prox_{\step f_j}$ for some stepsize $\step > 0$.  However, in
practice, these proximal operators will not evaluated exactly, so that
it is convenient to state the algorithm more generally in terms of
proximal solvers with the property that
\begin{align*}
\proxj{x} \approx \prox_{\step f_j}(x), \qquad \text{for all}~x \in
\R^d,
\end{align*}
for a suitably chosen stepsize $\step > 0$.  We make the sense of this
approximation precise in Section~\ref{sec:convergence-analysis}, where
we give convergence results under access to both exact and approximate
proximal oracles.

An immediate advantage to the scheme above is that it preserves the
correct fixed points for the distributed problem:

\bpr
\label{prop:fixed-point-ourmethod}
Given any $\step > 0$, suppose that $\zfp =
  (\zfp_1, \dots, \zfp_\numdevices)$ is a fixed point for
the \fedsplitspace \; procedure (Algorithm~\ref{alg:ourmethod}),
  meaning that
\begin{align}
  \zfp_j = \zfp_j + 2\left(\prox_{\step f_j}(2 \overline{\zfp} -
  \zfp_j) - \overline{\zfp}\right), \qquad \mbox{for all $j \in
    [\numdevices]$.}
\end{align}
Then the average $\xstar \coloneq \frac{1}{\numdevice}
\sum_{j=1}^\numdevices \zfp_j$ is an optimal solution to the
original distributed problem---that is,
\begin{align*}
  \sum_{j=1}^\numdevices f_j(\xstar) =
  \inf_{x \in \real^d} \sum_{j=1}^\numdevices f_j(x).
\end{align*}
\epr
\noindent See Section~\ref{SecProofCorrectFixPoint} for the proof of this claim. \\

Note that Proposition~\ref{prop:fixed-point-ourmethod} does not say
anything about the convergence of the \fedsplitspace scheme. Instead,
it merely guarantees that if the iterates of the method do converge,
then they converge to optimal solutions of the problem that is being
solved.  This is to be contrasted with
Propositions~\ref{prop:FedGD-fixed-points}
and~\ref{prop:FedProx-fixed-points}, that show the incorrectness of
other proposed algorithms.  It is the focus of the next section to
derive conditions under which we can guarantee convergence of the
\fedsplitspace scheme.

%%%%%%%%%%%%%%%%%%%%%%%%%%%%%%%%%%%%%%%%%%%%%%%%%%%%%%%%%%%%%%%%%%%%%%%%%%%%%%%%%%%%%%%%%%%%%%%%%%%%%%%%%%%%
\subsection{Convergence results}
\label{sec:convergence-analysis}

In this section, we give convergence guarantees for the \fedsplitspace
procedure in Algorithm~\ref{alg:ourmethod}.  By appealing to classical
first-order convex optimization theory, we are also able to give
iteration complexities under various proximal operator
implementations.

\subsubsection{Strongly convex and smooth losses}

We begin by considering the case when the losses $f_j \colon \R^d \to
\R$ are $\ell_j$-strongly convex and $L_j$-smooth.  We define the
quantities
\begin{align}
\label{EqnKeyQuantities}  
\ellmin \coloneq \min_{j=1,\dots,\numdevices} \ell_j, \quad \Lmax
\coloneq \max_{j=1,\dots,\numdevices} L_j, \quad \text{and} \quad
\kappa \coloneq \frac{\Lmax}{\ellmin}.
\end{align}
Note that $\ellmin$ corresponds to the smallest strong convexity
parameter; $\Lmax$ corresponds to the largest smoothness parameter;
and $\kappa$ corresponds to the induced condition number of such a
problem.

The following result demonstrates that in this setting, our method
enjoys geometric convergence to the optimum, even with inexact
proximal implementations.

\btheo
\label{thm:inexact-convergence}
Suppose that the local proximal updates of
Algorithm~\ref{alg:ourmethod} (Step 1A) are possibly inexact, with
errors bounded as
\begin{align}
\label{EqnUniformProxError}  
  \|\proxj{z} - \prox_{\step f_j}(z)\| \leq b \qquad \text{for
    all}~j~\text{and all}~z \in \R^d.
\end{align}
Then for any initialization $z^{(1)}\in \R^d$, the \fedsplitspace
algorithm with stepsize $\step = 1/\sqrt{\ellmin \Lmax}$
satisfies the bound
\begin{align}
\label{EqnFedSplitBound}
\|x^{(t+1)} - \xstar\| \leq \left(1 - \frac{2}{\sqrt{\kappa} +
  1}\right)^{\! t}\frac{\|z^{(1)} - \zstar\|}{\sqrt{m}} +
(\sqrt{\kappa} + 1)b, \qquad \mbox{for all $t = 1, 2, \ldots$.}
\end{align}
\etheo
\noindent We prove Theorem~\ref{thm:inexact-convergence} in
Section~\ref{sec:convergence} as a consequence of a more general
result that allows for different proximal evaluation error at each
round, as opposed to the uniform bound~\eqref{EqnUniformProxError}
assumed here.

%%%%%%%%%%%%%%%%%%%%%%%%%%%%%%%%%%%%%%%%%%%%%%%%%%%%%%%%%%%%%%%%%%%%%%%%%%%%%%%%%%%%%%%%

\paragraph{Exact proximal evaluations:}  In the special (albeit unrealistic) case
when the proximal evaluations are exact, the uniform
bound~\eqref{EqnUniformProxError} holds with $b = 0$, and the
bound~\eqref{EqnFedSplitBound} simplifies to
\begin{align*}
\|x^{(t+1)} - \xstar\| & \leq \left(1 - \frac{2}{\sqrt{\kappa} + 1}
\right)^{\! t} \frac{\|z^{(1)} - \zstar\|}{\sqrt{m}}.
\end{align*}
Consequently, given some initial vector $z^{(1})$, if we want to obtain
a solution $x^{(T+1)}$ that is $\epsilon$-accurate (i.e., with
$\|x^{(T)} - \xstar\| \leq \epsilon$), it suffices to take
\begin{align*}
T(\epsilon, \kappa) & = c \sqrt{\kappa} \log \left( \frac{\|z^{(1)} -
  \zstar\|}{\epsilon \sqrt{\numdevice}} \right)
\end{align*}
iterations of the overall procedure, where $c > 0$ is a universal
constant.

\paragraph{Approximate proximal updates by gradient steps:}  In practice,
the \fedsplitspace algorithm will be implemented using an approximate
prox-solver; here we consider doing so by using a gradient method on
each device $j$.  Recall that the proximal update at device $j$ at
round $t$ takes the form:
\begin{align*}
\prox_{\step f_j}(x_j^{(t)}) & = \argmin_{u \in \real^d} \big \{
\underbrace{\step f_j(u) + \frac{1}{2} \|u - x_j^{(t)}\|_2^2}_{h_j(u)}
\big \}.
\end{align*}
A natural way to compute an approximate minimizer is to run
$\numepoch$ rounds of gradient descent on the function $h_j$.  (To be
clear, this is \emph{not} the same as running multiple rounds of
gradient descent on $f_j$ as in the \texttt{FedGD} procedure.)
Concretely, at round $t$, we initialize the gradient method with the
initial point $u^{(1)} = x_j^{(t)}$, let us run gradient descent on
$h_j$ with a stepsize $\gstep$, thereby generating the sequence
\begin{align}
  u^{(t+1)} & = u^{(t)} - \gstep \nabla h_j(u^{(t)}) \; = \; u^{(t)} -
  \alpha \step \nabla f_j(u^{(t)}) + \big( u^{(t)} - x_j^{(t)} \big)
\end{align}
We define $\proxj{x_j^{(t)}}$ to be the output of this procedure after
$\numepoch$ steps.

\bcors[\fedsplitspace convergence with inexact proximal updates]
\label{CorFedSplitStrong}
Consider the \fedsplitspace \; procedure run with proximal stepsize
$\step = \frac{1}{\sqrt{\ellmin \Lmax}}$, and using approximate
proximal updates based on $\numepoch$ rounds of gradient descent with
stepsize \mbox{$\gstep = (1 + \step \tfrac{\ellmin + \Lmax}{2})^{-1}$}
initialized (in round $t$) at the previous iterate $x_j^{(t)}$.  Then
the the bound~\eqref{EqnUniformProxError} holds at round $t$ with
error at most
\begin{align}
\label{EqnProxErrorBound}
b & \leq \big(1 - \frac{1}{\sqrt{\kappa} + 1} \big)^\numepoch \; \|
x_j^{(t)} - \prox_{\step f_j}(x_j^{(t)})\|_2.
\end{align}
\ecors

Given the exponential decay in the number of rounds $\numepoch$
exhibited in the bound~\eqref{EqnProxErrorBound}, in practice, it
suffices to take a relatively small number of gradient steps.  For
instance, in our experiments to be reported in
Section~\ref{sec:numerics}, we find that $\numepoch = 10$ suffices to
track the evolution of the algorithm using exact proximal updates up
to relatively high precision.

\paragraph{Comments on stepsize choices:}  
It should be noted that the guarantees provided in
Theorem~\ref{thm:inexact-convergence} and
Corollary~\ref{CorFedSplitStrong} both depend on stepsize choices that
involve knowledge of the smoothness parameter $\Lmax$ and/or the
strong convexity parameter $\ellmin$, as defined in
equation~\eqref{EqnKeyQuantities}.  With reference to the gradient
updates in Corollary~\ref{CorFedSplitStrong}, we can adapt standard
theory (e.g.,~\cite{BoyVan04}) to show that if the gradient stepsize
parameter $\alpha$ were chosen with a backtracking line search, we
would obtain the same error bound~\eqref{EqnProxErrorBound}, up to a
multiplicative pre-factor applied to the term $\tfrac{1}{\sqrt{\kappa}
  + 1}$.  As for the proximal stepsize choice $\step$, we are not
currently aware of standard procedures for setting it that are
guaranteed to preserves the convergence bound of
Theorem~\ref{thm:inexact-convergence}.  However, we believe that this
should be possible, and this is an interesting direction for future
research.

%%%%%%%%%%%%%%%%%%%%%%%%%%%%%%%%%%%%%%%%%%%%%%%%%%%%%%%%%%%%%%%%%%%%%%%%%%%%%%%%%%%%%

\subsubsection{Smooth but not strongly convex losses}

We now consider the case when $f_j \colon \R^d \to \R$ are
$L_j$-smooth and convex, but not necessarily strongly convex.  Given
these assumptions, the consensus objective $F(z) =
\sum_{j=1}^\numdevices f_j(z_j)$ is a $\Lmax$-smooth function on the
product space $(\R^d)^m$.  So as to avoid degeneracies, we assume that
the federated objective $x \mapsto \sum_{j=1}^\numdevices f_j(x)$ is
bounded below, and achieves its minimum.

Our approach to solving such a problem is to apply the \fedsplitspace
procedure to a suitably regularized version of the original problem.
More precisely, given some initial vector $x^{(1)} \in \R^d$ and a
regularization parameter $\lambda > 0$, let us define the function
\begin{align}
\label{eqn:regularized-objective}
F_\lambda(z) \defn \sum_{j=1}^\numdevice \Big\{ f_j(z_j) +
\frac{\lambda}{2} \|z_j - x^{(1)}\|^2 \Big\}.
\end{align}
We see that $F_\lambda \colon (\R^d)^m \to \R$ is a $\lambda$-strongly
convex and $\Lmax_\lambda = (\Lmax + \lambda)$-smooth function.  The
next result shows that for any $\eps > 0$, minimizing the function
$F_\lambda$ up to an error of order $\eps$, using a carefully chosen
$\lambda$, yields an $\eps$-cost-suboptimal minimizer of the original
objective function $F$.
%%%%%%%%%
\btheo
\label{thm:reduction}
Given some $\lambda \in \Big(0, \tfrac{\eps}{m \|x^{(1)} - \xstar\|^2}
\Big)$ and any initialization $x^{(1)} \in \R^d$, suppose that we run
the
\fedsplitspace procedure (Algorithm~\ref{alg:ourmethod}) on the
regularized objective $F_\lambda$ using exact prox steps with stepsize
$\step = 1/\sqrt{\lambda \Lmax_\lambda}$. Then the \fedsplitspace
algorithm outputs a vector $\xhat \in \R^d$ satisfying $F(\xhat) -
\Fstar \leq \eps$ after at most
\begin{align}
  \label{EqnFedSplitWeak}
\tildeO{\sqrt{\frac{\Lmax \|x^{(1)} - \xstar\|^2}{\eps}}}
\end{align}
iterations.
\etheo
\noindent See Section~\ref{proof:reduction} for the proof of this
result. \\

\noindent To be clear, the $\tildeO{\cdot}$ notation in the
bound~\eqref{EqnFedSplitWeak} denotes the presence of constant and
polylogarithmic factors that are not dominant.

%%%%%%%%%%%%%%%%%%%%%%%%%%%%%%%%%%%%%%%%%%%%%%%%%%%%%%%%%%%%%%%%%%%%%%%%%%%%%%%%%%%%%%%%%%%%%

\section{Experiments}
\label{sec:numerics}

In this section, we present some simple numerical results for
\fedsplit.  We begin by presenting results for least squares problems
in Section~\ref{sec:exp-leastsquares} as well as logistic regression
in Section~\ref{sec:exp-logreg}.  This section concludes with a
comparison of the performance of \fedsplitspace versus federated
gradient procedures in Section~\ref{sec:exp-comparison}.  All of the
experiments here were conducted on a machine running Mac OS 10.14.5,
with a 2.6 GHz Intel Core i7 processor, in \texttt{Python} 3.7.3.  In
order to implement the proximal operators for the logistic regression
experiments, we used \texttt{CVXPY}, a modelling language for
disciplined convex programs~\cite{DiaBoy16}.

\subsection{Least squares}
\label{sec:exp-leastsquares}

As an initial object of study, we consider the least squares
problem~\eqref{prob:least-squares}.  In order to have full control
over the conditioning of the problem, we consider instances defined by
randomly generated datasets.   Given a collection of design matrices
$A_j$ for $j \in [\numdevices]$, we generate random response vectors
according to a linear measurement model
\begin{align*}
b_j = A_j \xtrue + \noisevec_j, \quad \mbox{for $j \in [\numdevices] =
  \{1, \ldots, \numdevices\}$,}
\end{align*}
where the noise vector is generated as $\noisevec_j \simind
\Normal{0}{\sigma^2I_{n_j}}$.
For this experiment, we set
\begin{align*}
d = 500, \quad m = 25, \quad n_j \equiv 5000, \quad \text{and} \quad
\sigma^2 = 0.25.
\end{align*}
We also generate random versions of the design matrices $A_j$, from
one of two possible ensembles:
\begin{itemize}
\item Isotropic ensemble: each design matrix $A_j \in \real^{\numobs_j
  \times \usedim}$ is generated with \iid entries $(A_j)_{k \ell}
  \simiid \Normal{0}{1}$, for all $k \in [n_j]$ and $\ell \in [d]$.
  In the regime $n_j \gg \usedim$ considered here, known results in
  non-asymptotic random matrix theory (e.g.~\cite{Wai19}) guarantee
  that the matrix $A_j^\T A_j$ will be well-conditioned.
\item Spiked ensemble: in order to illustrate how algorithms depend on
  conditioning, we also generate design matrices $A_j$ according the
  procedure described in Section~\ref{sec:exp-comparison} with $\kappa
  = 10$.  This leads to a problem that has condition number $\kappa$.
\end{itemize}
Finally, we construct a random parameter vector $\xtrue$ by sampling
from the standard Gaussian distribution $\Normal{0}{I_d}$.

We solve this federated problem via \fedsplit, implemented with both
exact proximal updates as well as with a constant number of local
gradient steps, $\numepochs \in \{1, 5, 10\}$.  For comparison, we
also apply the \texttt{FedGD} procedure~\eqref{eqn:FedGD}.
\begin{figure}[h!]
  \begin{center}
    \begin{tabular}{cc}
          \widgraph{0.49\linewidth}{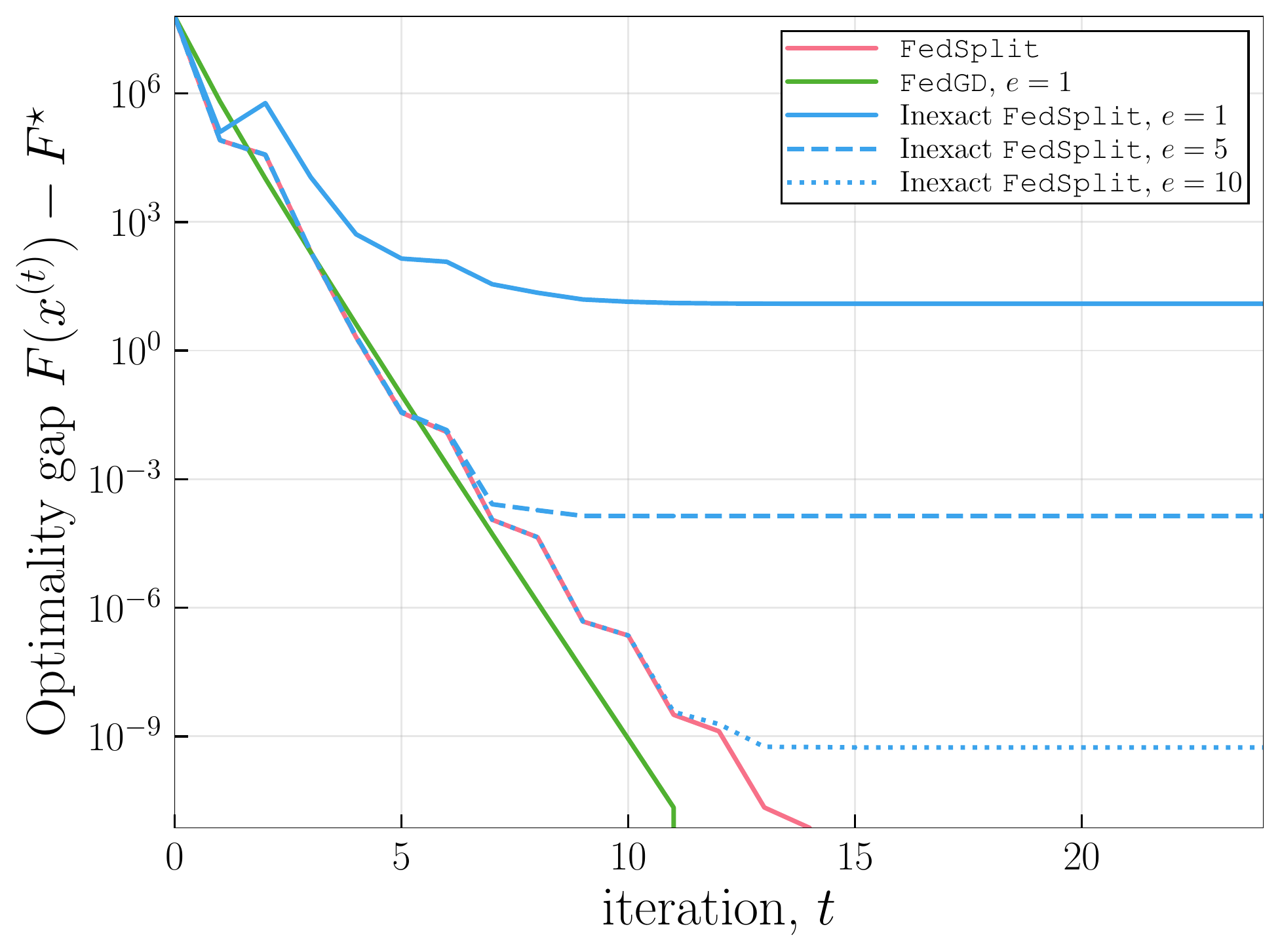} &
          \widgraph{0.49\linewidth}{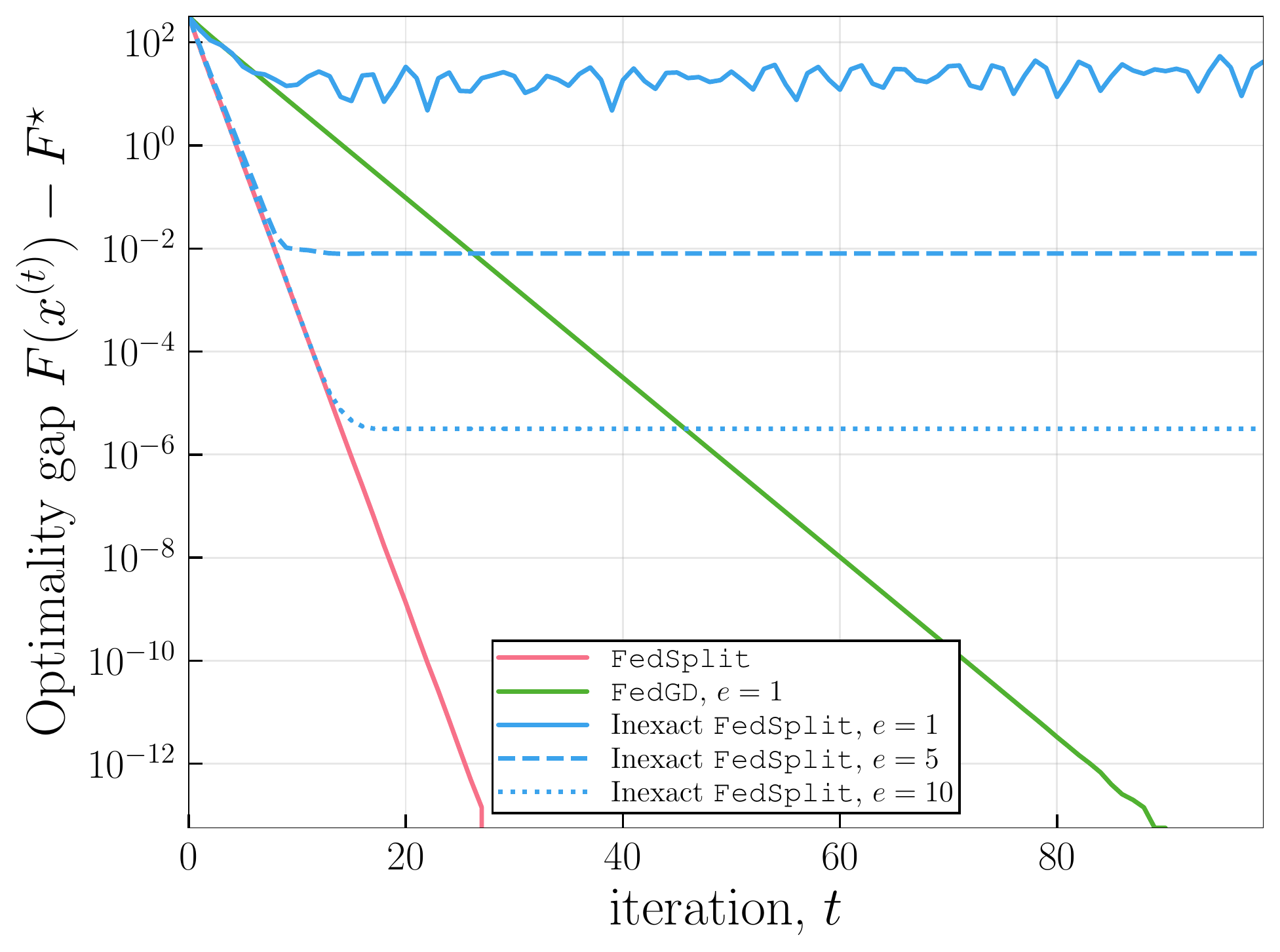} \\
          (a) Isotropic case & (b) Spiked case
    \end{tabular}
  \caption{Plots of the optimality gap $F(x^{(t)}) - \Fstar$ versus
    the iteration number $t$ for various algorithms as applied to a
    simple least-squares problem~\eqref{prob:least-squares}.  Here
    $\Fstar$ denotes the optimal cost value and $F(x^{(t)})$ denote
    the cost returned at round $t$ by a given algorithm.  Shown are
    curves for inexact and exact implementations of
    the \fedsplitspace$\,$ method and the federated gradient method
    with $\numepochs = 1$.  The inexact implementations of
    the \fedsplitspace$\,$ method use a gradient method for the
    approximate proximal updates, and we show curves for number of
    gradient steps $\numepoch \in \{1, 5, 10\}$.  These curves exhibit
    floors corresponding to the errors introduced by the approximate
    proximal solves. (a) Isotropic case: results for random design
    matrices with \iid~$\Normal{0}{I_d}$ rows, corresponding to a
    well-conditioned setting. (b) Spiked case: results for design
    matrices generated according to the procedure described in
    Section~\ref{sec:exp-comparison} with $\kappa = 10$.  These
    problems are somewhat poorly conditioned, causing a slowdown in
    convergence that is more severe for federated gradient.}
  \label{fig:leastsquares}  
  \end{center}
\end{figure}
We solve the resulting optimization problem using various methods: the
\texttt{FedGD} procedure with $\numepoch = 1$, which is the only
setting guaranteed to preserve the fixed points; the exact form
of \fedsplitspace procedure, in which the proximal updates are
computed exactly; and inexact versions of the \fedsplitspace procedure
using the gradient method (see Corollary~\ref{CorFedSplitStrong}) to
compute approximations to the updates with $\numepoch \in \{1, 5,
10\}$ rounds of gradient updates per machine.

Figure~\ref{fig:leastsquares} shows the results of these experiments,
plotting the log optimality gap $\log (F(x^{(t)}) - F^*)$ versus the
iteration number $t$ for these algorithms; see the caption for
discussion of the behavior.

%%%%%%%%%%%%%%%%%%%%%%%%%%%%%%%%%%%%%%%%%%%%%%%%%%%%%%%%%%%%%%%%%%%%%%%%%%%%%%%%%%%

\subsection{Logistic regression}
\label{sec:exp-logreg}

Moving beyond the setting of least squares regression, we now explore
the behavior of various algorithms for solving federated binary
classification.  In this problem, we again have fixed design matrices
$A_j \in \R^{n_j \times d}$, but the response vectors take the form of
labels, $b_j \in \{1, -1\}^{n_j}$.  Here the rows of $A_j$, denoted by
$a_{ij} \in \R^d$ for $i = 1, \dots, n_j$ are collections of $d$
features, associated with class label $b_{ij} \in \{-1, 1\}$, the
entries of $b_j$.  We assume that for $j = 1,\dots,\numdevices$ and
unknown parameter vector $\xtrue \in \real^d$, the conditional
probability of observing a positive class label $b_{ij} = 1$ is given
by
\begin{align}
\label{EqnBernoulli}
\P\{b_{ij} = 1\} = \frac{\e^{a_{ij}^\T \xtrue}}{1 + \e^{a_{ij}^\T
    \xtrue}}, \quad \text{for } i ~ =1,\dots, n_j.
\end{align}
Given observations of this form, the maximum likelihood estimate for
$\xtrue$ is then a solution to the convex program
\begin{align}
  \label{prob:logreg}
\text{minimize}\quad  \sum_{j=1}^\numdevices \sum_{i=1}^{n_j}
\log(1 + \e^{-b_{ij} a_{ij}^\T x}),
\end{align}
with variable $x\in \R^d$.
This problem is referred to as \emph{logistic regression}.

Since the function $t \mapsto \log(1 + \e^{-t})$ has bounded, positive
second derivative, it is straightforward to verify that the objective
function in problem~\eqref{prob:logreg} is smooth and convex.  The
local cost functions $f_j \colon \R^d \to \R$ are given by the
corresponding sums over the local datasets---that is
\begin{align*}
f_j(x) \defn \sum_{i=1}^{n_j} \log(1 + \e^{-b_{ij} a_{ij}^\T x}),
\qquad \text{for}~j = 1,\dots, \numdevices.
\end{align*}
With this definition, we see that~\eqref{prob:logreg} is equivalent to
the minimization of $F(x) = \sum_{j=1}^\numdevices f_j(x)$, which
places this problem in the broader framework of federated problems of
the form~\eqref{prob:finite-sum-problem}.

For the simple experiments reported here, we construct some random
instances of logistic regression problems with the settings
\begin{align*}
d = 100, \quad n_j \equiv 1000, \quad \text{and} \quad m = 10,
\end{align*}
so that the total sample size is $n = 10000$.  We construct a random
instance by drawing the feature vectors $a_{ij} \simiid
\Normal{0}{I_d}$ for $j = 1, \dots, \numdevices$ and $i = 1, \dots,
n_j$, generating the true parameter randomly as $\xtrue \simiid
\Normal{0}{I_d}$, and then generating the binary labels $b_{ij}$
according to the Bernoulli model~\eqref{EqnBernoulli}.

Given this synthetic dataset, we then solve the
problem~\eqref{prob:logreg} by applying exact implementations
of \fedsplitspace \; as well as a number of inexact implementations,
where each local update is a constant number of gradient steps,
$\numepochs \in \{1, 5, 10\}$.  For comparison, we also implement a
federated gradient method as previously described~\eqref{eqn:FedGD}.
\begin{figure}[h]
\begin{center}
\widgraph{0.65\linewidth}{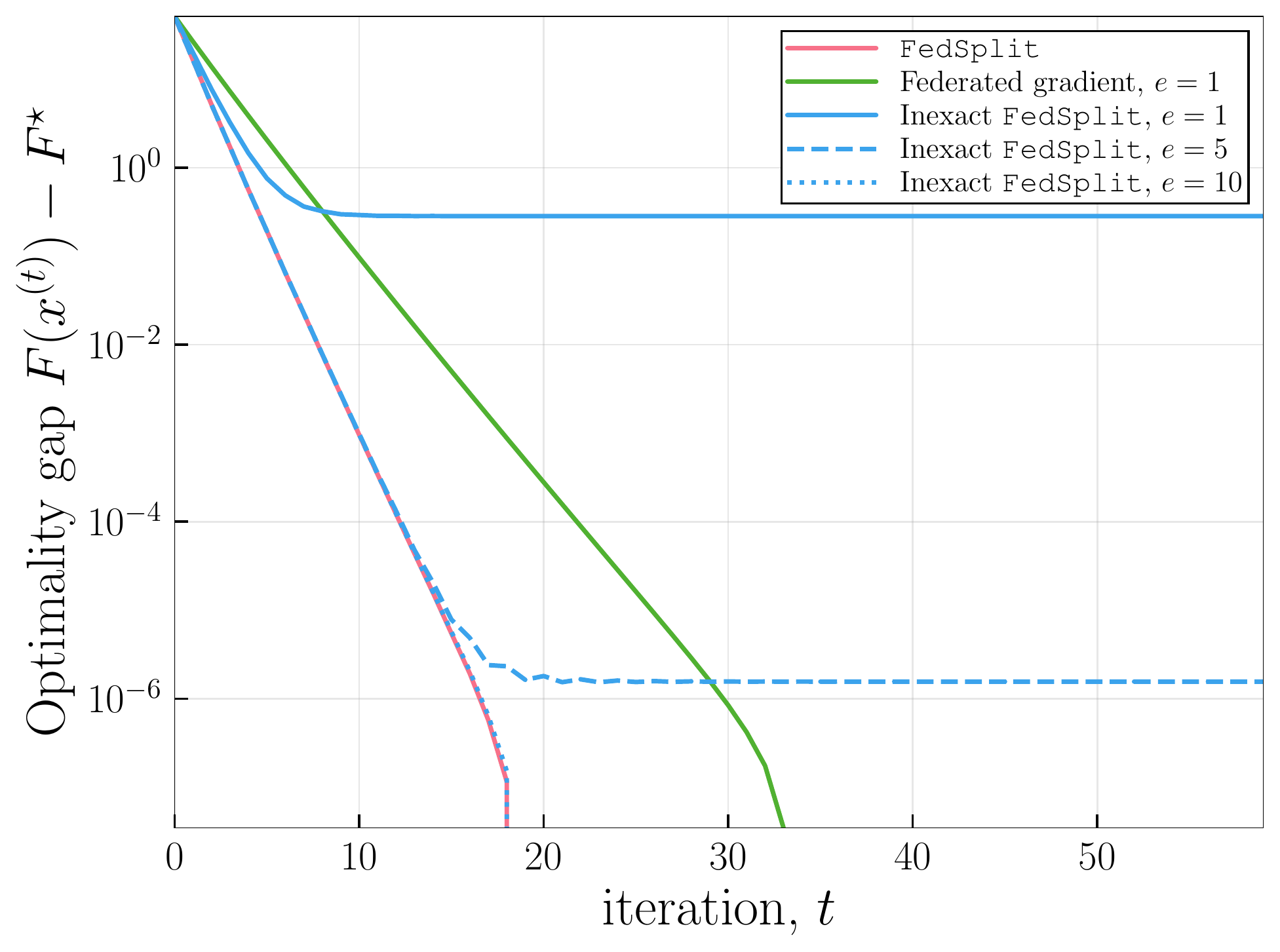}
\caption{ Plots of the optimality gap $F(x^{(t)}) - \Fstar$ versus the
  iteration number $t$ for various algorithms as applied to a logistic
  regression problem~\eqref{prob:logreg}.  Here $\Fstar$ denotes the
  optimal cost value and $F(x^{(t)})$ denote the cost returned at
  round $t$ by a given algorithm.  Shown are curves for inexact and
  exact implementations of the \fedsplitspace method and the federated
  gradient method with $\numepochs = 1$.  The inexact implementations
  of the \fedsplitspace method use a gradient method for the
  approximate proximal updates, and we show curves for number of
  gradient steps $\numepoch \in \{1, 5, 10\}$.  These curves for
  $\numepoch \in \{1, 5 \}$ exhibit an error floor (corresponding to
  the inaccuracy introduced by approximate proximal solves), whereas
  the curve for $\numepoch = 10$ tracks the exact procedure to
  accuracy below $10^{-6}$. }
  \label{fig:logreg}
\end{center}
\end{figure}
As shown in Figure~\ref{fig:logreg}, the results are qualitatively
similar to those shown for the least-squares problem.  Both
\texttt{FedGD} with $\numepoch = 1$ and the \fedsplitspace procedure
exhibit linear convergence rates.  Using inexact proximal updates with the
\fedsplitspace procedure preserves the linear convergence up to the
error floor introduced by the exactness of the updates.  In this case,
the inexact proximal updates with $\numepoch = 10$---that is,
performing $10$ local updates per each round of global
communication---suffice to track the exact
\fedsplitspace procedure up to an accuracy below $10^{-6}$.

%%%%%%%%%%%%%%%%%%%%%%%%%%%%%%%%%%%%%%%%%%%%%%%%%%%%%%%%%%%%%%%%%%%%%%%%%%%%%%%%%%%

\subsection{Dependence on problem conditioning}
\label{sec:exp-comparison}

It is well-known that the convergence rates of various optimization
algorithms can be strongly affected by the conditioning of the
problem, and theory makes specific predictions about this dependence,
both for the correct form of the \texttt{FedGD} algorithm (implemented
with $\numepoch = 1$), and the
\fedsplitspace procedure.  In practice, for the machine learning
prediction problems, ill-conditioning can arise due to heterogeneous
scalings and/or dependencies between different features that are used.
Accordingly, it is interesting to study the dependence of procedures on
the condition number.

First, let us re-state the theoretical guarantees that are enjoyed by
the different procedures.  For a given algorithm, its \emph{iteration
  complexity} is a measure of the number of iterations, as a function
of some error tolerance
$\eps> 0$ and possibly other problem
parameters, required to return a solution that is $\eps$-optimal.
More precisely, for a given algorithm, we let $T(\eps, \kappa)$ denote
the maximum number of iterations required so that, for any problem
with condition number at most $\kappa$, the iterate $x^{T}$ with $T =
T(\eps, \kappa)$ satisfies the bound $F(x^{(T)})- \Fstar \leq \eps$.
Note that in the federated setting, $T(\eps, \kappa)$ provides an
upper bound on the number of communication rounds required to ensure
an $\eps$-cost-suboptimal point to the federated optimization problem
that is being solved. Since communication is very often the dominant cost
in carrying out such numerical procedures, it is of great interest to make
$T(\eps, \kappa)$ as small as possible.

As mentioned in Section~\ref{sec:fedgrad}, classical results guarantee
that the federated gradient procedure~\eqref{eqn:FedGD}, when
implemented with the number of gradient steps $\numepochs = 1$, enjoys
linear convergence.  More precisely, given the federated objective
$F(z) = \sum_{j=1}^\numdevices f_j(z_j)$, we define its condition number
as $\kappa = \Lmax/\ellmin$.  Then \texttt{FedGD} has an iteration
complexity bounded as
\begin{subequations}
  \begin{align}
\Tfedgrad(\eps, \kappa) = O(\kappa \log(1/\eps)).
\end{align}
On the other hand, it follows from
Theorem~\ref{thm:inexact-convergence} that the \fedsplitspace
procedure has iteration complexity scaling as
\begin{align}
\Tfedsplit(\eps, \kappa) = O(\sqrt{\kappa}\log(1/\eps)).
\end{align}
\end{subequations}
Thus, albeit at the expense of a more expensive local update,
the \fedsplitspace \; procedure has better dependence on the condition
number than federated gradient descent.  This highlights an important
tradeoff between local computation and global communication in these
methods.

We now describe the results of a simulation study that demonstrates
the accuracy of these predicted iteration complexities.  At a high
level, our strategy is to construct a sequence of problems, indexed by
an increasing sequence of condition numbers $\kappa$, and to estimate
the number of iterations required to achieve a given tolerance $\eps >
0$ as a function of $\kappa$.  In order to do, it suffices to consider
ensembles of least squares problems~\eqref{prob:least-squares}, but
with a carefully constructed collection of design matrices, which
we now describe.

For a given integer $\ell \geq 2$, let $\Orthog(\ell)$ denote the set
of $\ell \times \ell$ orthogonal matrices over the reals, and let
$\Unif(\Orthog(\ell))$ denote the uniform (Haar) measure on this
compact group.  With this notation, we begin by sampling \iid random
matrices
\begin{align}
U_j^{(\kappa)} \sim \Unif(\Orthog(n_j)) \quad \text{and} \quad
V_j^{(\kappa)} \sim \Unif(\Orthog(d)), \qquad \text{for $j = 1,
  \ldots, \numdevices$.}
\end{align}
For a given condition number $\kappa \geq 1$, we define a padded
diagonal matrix---that is
\begin{align*}
\Lambda_j^{(\kappa)} = \begin{bmatrix} \diag(\lambda_j^{(\kappa)}) &
  0_{d, (n-d)} \end{bmatrix} \quad \text{where} \quad
\lambda_j^{(\kappa)} = (\sqrt{\kappa}, 1, \dots, 1) \in \R^d.
\end{align*}
Above, the matrix $0_{d, (n_j-d)} \in \R^{d \times (n_j-d)}$ has all
entries equal to zero.  Given the random orthogonal matrices and the
matrix $\Lambda_j^{(\kappa)} \in \R^{n_j \times d}$, we then construct
the design matrices $A_j^{(\kappa)} \in \R^{n_j \times d}$ by setting
\begin{align*}
A_j^{(\kappa)} \defn U_j^{(\kappa)} \Lambda_j^{(\kappa)}
V_j^{(\kappa)}, \quad \text{for all }~j=1,\dots, \numdevices.
\end{align*}
These choices ensure that the federated least squares
objective~\eqref{prob:least-squares} has condition number $\kappa$.

As before, the response vectors $b_j^{(\kappa)}$ obey a Gaussian
linear measurement model,
\begin{align*}
b_j^{(\kappa)} = A_j^{(\kappa)} x_0 + \noisevec_j^{(\kappa)}, \quad
\text{for}~j=1,\dots,\numdevices, \quad \text{and for all}~\kappa \in
K.
\end{align*}
We again take $\noisevec_j^{(\kappa)} \simind
\Normal{0}{\sigma^2I_{n_j}}$.  In our experiments, we draw the
parameter $x_0 \sim \Normal{0}{I_d}$, and use the parameter settings
\begin{align*}
\numdevice = 10, \quad d = 100, \quad n_j \equiv 400, \quad \mbox{and}
\quad \sigma^2 = 1.
\end{align*}
With these settings, we iterated over a collection of condition
numbers $\kappa \in \{10^0, 10^{0.5}, \dots, 10^{3.5}, 10^{4}\}$.  For
each choice of $\kappa$, after generating a random instance as
described above, we measured the number of iterations required for
\texttt{FedGD} and the \fedsplitspace procedures, respectively, to
reach a target accuracy $\eps = 10^{-3}$, which is modest at
best. 

\begin{figure}[h!]
\begin{center}
  \subfloat[]{{\widgraph{0.48\linewidth}{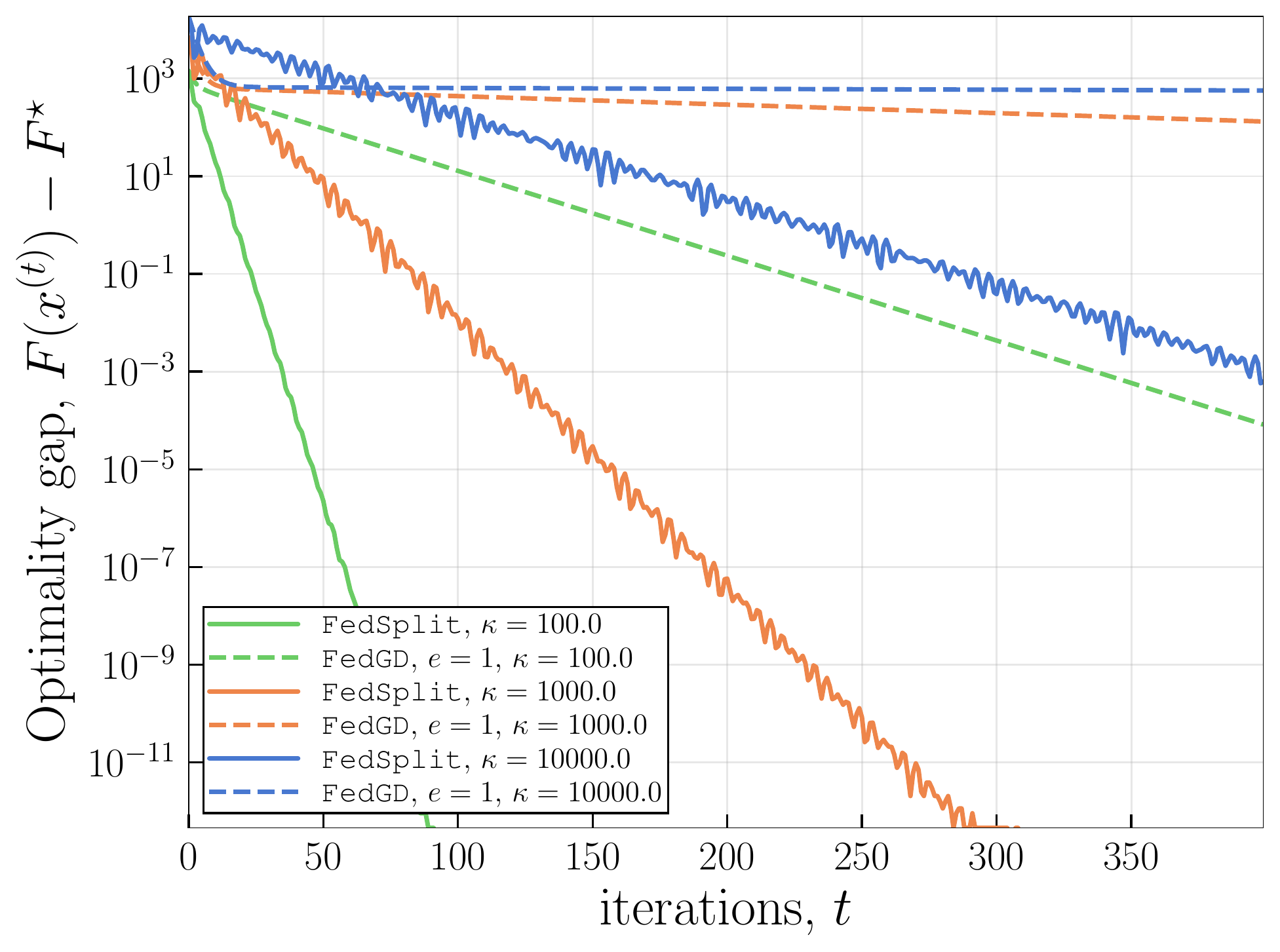}}}
  \hfill
    \subfloat[]{{\widgraph{0.48\linewidth}{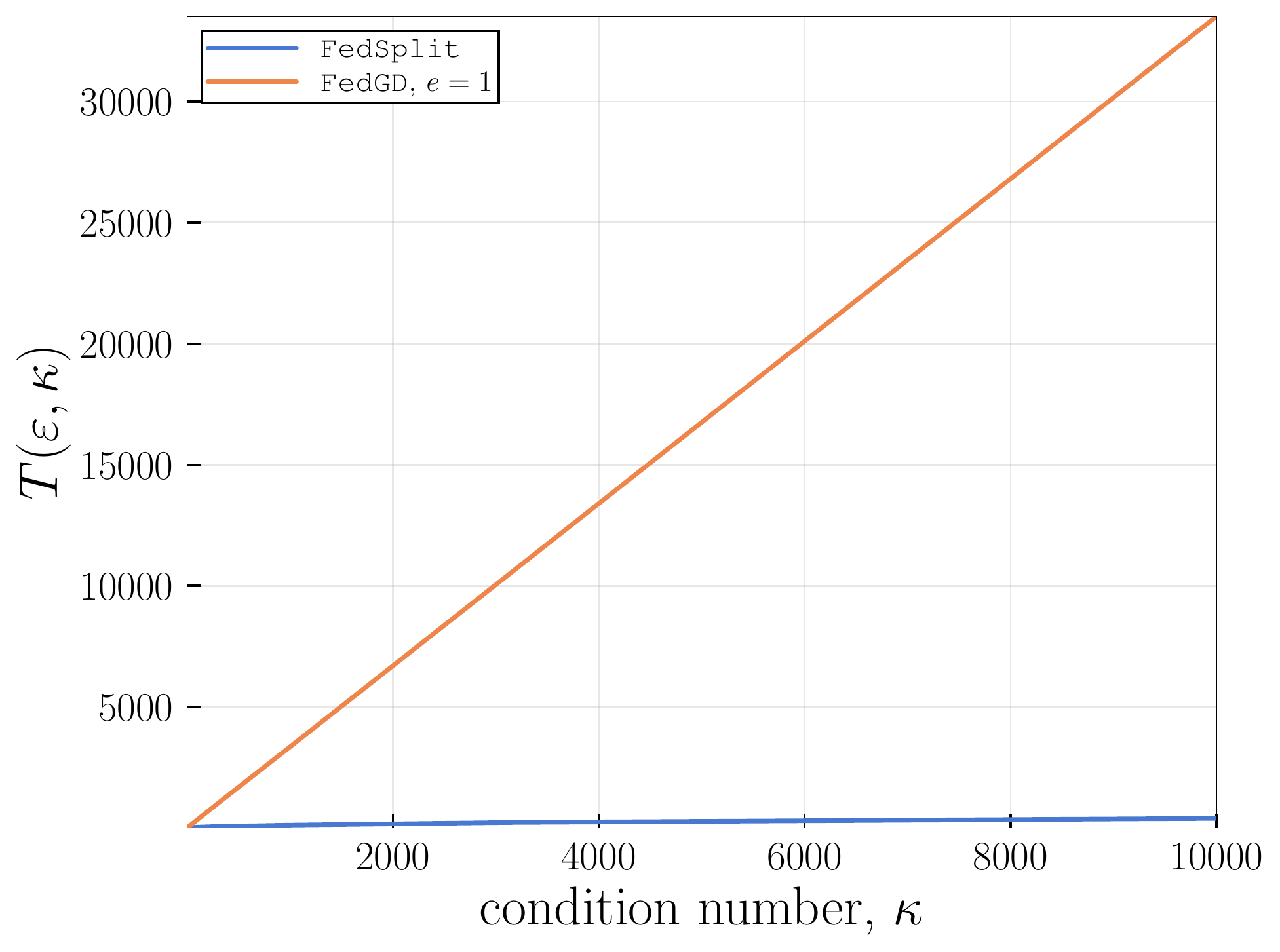}}}
  \caption{Dependence of algorithms on the conditioning. (a) Plot of
    log cost suboptimality of iterate $x^{(t)}$ versus iteration $t$
    for condition number $\kappa \in \{100, 1000, 10000\}$. Consistent
    with theory, both procedures exhibit geometric convergence, with
    the rates decaying as the condition number $\kappa$ increases.
    Note that the \texttt{FedGD} procedure degrades more rapidly than
    the \fedsplitspace procedure as $\kappa$ is increased.  (b) Plots
    of the iteration complexity $T(\eps; \kappa)$ versus $\kappa$ at
    tolerance level $\eps = 10^{-3}$ for the \texttt{FedGD}
    and \fedsplitspace procedures.  It measures the number of
    iterations required to achieve an $\eps = 10^{-3}$ accurate
    solution for a problem with condition number $\kappa$. Consistent
    with their theoretical guarantees, \texttt{FedGD} exhibits linear
    dependence, whereas the \fedsplitspace procedure exhibits
    sublinear dependence.}
    
  \label{fig:contraction}
\end{center}
\end{figure}

In this way, we obtain estimates of the functions $\kappa \mapsto
\Tfedgrad(10^{-3}, \kappa)$ and $\kappa \mapsto \Tfedsplit(10^{-3},
\kappa)$, which measure the dependence of the iteration complexity on
the condition number.  Figure~\ref{fig:contraction} provides plots of
these estimated functions.  Consistent with the theory, we see that
\texttt{FedGD} has an approximately linear dependence on the
condition number, whereas the \fedsplitspace procedure has much milder
dependence on conditioning.  Concretely, for an instance with
condition number $\kappa = 10000$, the \texttt{FedGD} procedure
requires on the order of $34000$ iterations, whereas the \fedsplitspace
procedure requires roughly $400$ iterations.  Again, to be fair,
the \fedsplitspace involves more complicated proximal updates at each
client, so that these iteration counts should be viewed as
reflecting the number of rounds of communication between the
individual clients and the centralized server, as opposed to total
computational complexity.

%%%%%%%%%%%%%%%%%%%%%%%%%%%%%%%%%%%%%%%%%%%%%%%%%%%%%%%%%%%%%%%%%%%%%%%%%%%%%%%%%%%%%%%%%%

\section{Proofs}
\label{SecProofs}

We now turn to the proofs of our main results.  Prior to diving into
these arguments, we first introduce two operators that play a critical
role in our analysis. Given a convex function $\phi \colon \R^d \to
\R$, we define
\begin{subequations}
\begin{align}
\label{EqnProxOperator}
\prox_{\phi}(z) & \coloneq \argmin_{x \in \R^d} \left \{ \phi(x) +
\frac{1}{2} \|z - x\|^2 \right \} \quad \text{and} \quad \\
\refl_{\phi}(z) & \coloneq 2 \prox_\phi(z) - z.
\end{align}
\end{subequations}
These are called the proximal and reflected resolvent operators
associated with the function $\phi$.  The first operator is also known
as the resolvent; the second operator above is also known as the
Cayley operator of $\phi$. Moreover, our analysis makes use of the
(semi)norm on Lipschitz continuous functions $f \colon \R^d \to \R$
given by
\begin{align}
\Lip(f) \coloneq \sup_{x \neq y} \frac{|f(x) - f(y)|}{\|x - y\|}.
\end{align}
For short, we say that that $f$ is $\Lip(f)$-Lipschitz continuous when
it satisfies this condition.

%%%%%%%%%%%%%%%%%%%%%%%%%%%%%%%%%%%%%%%%%%%%%%%%%%%%%%%%%%%%%%%%%%%%%%%%%%%%%%%%%%%%%%%%%%%%%%%%%%%%%%

\subsection{Proofs of guarantees for $\fedsplit$}
\label{sec:convergence}

We begin by proving our guarantees for the \fedsplitspace procedure,
including the correctness of its fixed points
(Proposition~\ref{prop:fixed-point-ourmethod}); the general
convergence guarantee in the strongly convex case
(Theorem~\ref{thm:inexact-convergence}); the general convergence
guarantee in the weakly convex case (Theorem~\ref{thm:reduction}), and
Corollary~\ref{CorFedSplitStrong} on its convergence with approximate
proximal updates.

\subsection{Proof of Proposition~\ref{prop:fixed-point-ourmethod}}
\label{SecProofCorrectFixPoint}

By the fixed point assumption, the block average $\xstar \defn
\overline{\zfp}$
satisfies the relation
\begin{align*}
\prox_{\step f_j}(2 \xstar - \zfp_j) = \xstar \qquad \mbox{for $j
  = 1, 2, \ldots, \numdevices$.}
\end{align*}
Since each $f_j$ is convex and differentiable, by the first-order
stationary conditions implied by the definition of the prox
operator~\eqref{EqnProxOperator}, we must have
\begin{align*}
  \nabla f_j(\xstar) + \tfrac{1}{\step} \big \{ \xstar - \big(2 \xstar -
  \zfp_j) \big \} \; = \; \nabla f_j(\xstar) + \tfrac{1}{\step} \big \{
  \zfp_j - \xstar \big \} \; = \; 0 \quad \mbox{for $j = 1,
    \ldots, \numdevices$.}
\end{align*}
Summing these equality relations over $j = 1, \ldots, \numdevices$ and
using the fact that
$\xstar = \tfrac{1}{\numdevices} \sum_{j=1}^\numdevices \zfp_j$
yields the zero gradient condition
\begin{align*}
\sum_{j=1}^\numdevices \nabla f_j(\xstar) \; = \; 0.
\end{align*}
Since the function $x \mapsto \sum_{j=1}^\numdevices f_j(x)$ is
convex, this zero-gradient condition implies that $\xstar \in \real^d$
is a minimizer of the distributed problem as claimed.

\subsubsection{Proof of Theorem~\ref{thm:inexact-convergence}}
We now turn to the proof of Theorem~\ref{thm:inexact-convergence}.
Our strategy is to prove it as a consequence of a somewhat more
general result, which we begin by stating here.  In order to lighten
notation, we use the fact that the proximal operator for the function
$F(z_1, \ldots, z_\numdevices) = \sum_{j=1}^\numdevices f_j(z_j)$ is
block-separable, so that in terms of the block-partitioned vector $z =
(z_1, \ldots, z_\numdevices)$, we can write
\begin{align*}
\prox_{\step F}(z) = \left(\prox_{\step f_1}(z_1), \dots, \prox_{\step
  f_\numdevices}(z_\numdevices) \right), \quad \mbox{for all $z =
  (z_1, \dots, z_\numdevices) \in (\real^d)^\numdevices$.}
\end{align*}
We also recall the the approximate proximal operator used in
the \fedsplitspace procedure, namely
\begin{align*}
\aprox(z) \coloneq \left(\aproxj{1}{z_1}, \dots,
\aproxj{\numdevices}{z_\numdevices}\right), \quad \text{for all}~z_1,
\dots, z_\numdevices \in \R^{d}.
\end{align*}

\btheo[Convergence with general residuals]
\label{thm:master-convergence}
Suppose that the functions $f_j \colon \R^d \to \R$ are
$\ell_j$-strongly convex and $L_j$-smooth for $j = 1, \ldots,
\numdevice$, and for $t = 1, 2, \ldots$, define the residuals
\begin{align}
  r^{(t)} \coloneq \aprox(2\overline{z^{(t)}} - z^{(t)}) -
  \prox_{\step F}(2\overline{z^{(t)}} - z^{(t)}).
\end{align}
Then with stepsize $\step = 1/\sqrt{\ellmin
    \Lmax}$, the \fedsplitspace procedure
  (Algorithm~\ref{alg:ourmethod}) has a unique fixed point $\zfp$, and
  the iterates satisfy
\begin{align}
  \|z^{(t + 1)} - \zfp \| & \leq \rho^{t}\|z^{(1)} - \zfp\| + 2
  \sum_{j=1}^t \rho^{t-j} \|r^{(j)}\| \qquad \mbox{for $t = 1, 2,
    \ldots$,}
\end{align}
where $\rho \coloneq 1 - 2/(\sqrt{\kappa} + 1)$ is the
contraction coefficient.  \etheo

Let us use Theorem~\ref{thm:master-convergence} to derive the claim
stated in Theorem~\ref{thm:inexact-convergence}.  Note that by
Proposition~\ref{prop:fixed-point-ourmethod}, the fixed points of
Algorithm~\ref{alg:ourmethod} are minimizers of $F$, hence unique
under the strong convexity assumption. Consequently, we have
\begin{align*}
\|x^{(t + 1)} - \xstar\| \leq \frac{1}{\sqrt{m}} \|z^{(t + 1)} -
\zstar\|, \qquad \mbox{for all $t = 1, 2, \ldots$.}
\end{align*}
Using Theorem~\ref{thm:master-convergence} and the error bound, we
then conclude that
\begin{align*}
\|x^{(t + 1)} - \xstar\| \leq \frac{1}{\sqrt{m}} \left(1 -
\frac{2}{\sqrt{\kappa} + 1} \right)^{\! t} \|z^{(1)} - \zstar\| +
(\sqrt{\kappa} + 1) b,
\end{align*}
as claimed.

%%%%%%%%%%%%%%%%%%%%%%%%%%%%%%%%%%%%%%%%%%%%%%%%%%%%%%%%%%%%%%%%%%%%%%%%%%%%%%%%%%%%%%%%%%%%%%%%%%%%%%%%%%%
  
\subsubsection{Proof of Theorem~\ref{thm:master-convergence}}

We now turn to the proof of the more general claim.  Given additive
decomposition \mbox{$F(z) = \sum_{j=1}^\numdevices f_j(z_j)$,} the
reflected resolvent induced by $F$ is block-separable, taking the form
\begin{align*}
\refl_{\step F}(z) = \left(\refl_{\step f_1}(z_1), \dots, \refl_{\step
  f_\numdevices}(z_\numdevices)\right), \quad \mbox{for all $z = (z_1,
  \dots, z_\numdevices) \in (\R^{d})^\numdevices$.}
\end{align*}
Similarly, consider the approximate reflected resolvent defined by the
algorithm, namely
\begin{align*}
\arefl(z) \coloneq 2 \aprox(z) - z, \quad \mbox{for all $z = (z_1,
  \ldots, z_\numdevices) \in (\R^{d})^\numdevices$.}
\end{align*}
It also has the same block-separable form.  

Using these two block-separable operators, we can now define two
abstract operators, each acting on the product space
$(\real^\usedim)^\numdevice$, that allow us to analyze the algorithm.
The first operator $\cT$ underlies the idealized algorithm, in which
the proximal updates are exact, and the second operator $\That$
underlies the practical algorithm, which is based on approximate
proximal updates.  The idealized algorithm is based on iterating the
operator
\begin{align}
  \label{EqnIdealized}
\cT(z) \coloneq \refl_{\step F} \big(
\refl_{I_\eqset}(z) \big).
\end{align}
In this definition, we use $I_\eqset$ to denote the indicator function
for membership in the equality subspace $\eqset$, so that $\refl_{I_\eqset}$
is the reflected proximal operator for this function.

On the other hand, the practical algorithm generates the sequence
$\{z^{(t)}\}_{t=1}^\infty$ via the updates $z^{(t+1)} =
\That(z^{(t)})$, where $\That: (\real^\usedim)^\numdevices \rightarrow
(\real^\usedim)^\numdevices$ is the \emph{perturbed operator}
\begin{align}
  \label{EqnPractical}
\That(z) & =  \arefl \big(\refl_{I_\eqset}(z)
\big).
\end{align}
Note that the idealized operator $\cT$ and perturbed operator $\That$
satisfy the relation
\begin{align}
  \label{EqnPerturb}
  \hat \cT - \cT & = \left(\arefl \circ \refl_{I_\eqset} -
  \refl_{\step F} \circ \refl_{I_\eqset}\right).
\end{align}
Our proof involves verifying that with the stepsize choice $\step =
1/\sqrt{\ellmin \Lmax}$, the mapping $\cT$ is a contraction, with
Lipschitz coefficient
\begin{align}
 \label{ineq:lip-factor}
\Lip(\cT) & \leq \underbrace{1 - \frac{2}{\sqrt{\kappa} + 1}}_{
  \mbox{$\eqcolon \rho$}} < 1.
\end{align}
Taking this claim as given for the moment, the contractivity implies
that $\cT$ has has a unique fixed point~\cite{GoeKir90}---call it
$\zstar \in (\real^\usedim)^\numdevices$.  Comparing with
Proposition~\ref{prop:fixed-point-ourmethod}, we see that the
definition of fixed points given there agrees with the fixed point
$\zstar$ of the operator $\cT$, since we have the relation
$\refl_{I_\eqset}(z) = 2\overline{z} - z$.
  
Using this contractivity condition, the distance between this fixed
point $\zstar$ and the iterates $z^{(t)}$ of the \fedsplitspace
procedure can be bounded as
\begin{align}
  \|z^{(t+1)} - \zfp\| &= \|\hat \cT z^{(t)} - \cT \zfp \| \nonumber \\
& \stackrel{\rm (i)}{\leq} \|\cT z^{(t)} - \cT \zfp \| + 2 \| \aprox
  \refl_{I_\eqset} z^{(t)} - \prox_{\step F} \refl_{I_\eqset} z^{(t)}
  \| \nonumber \\
& \stackrel{\rm (ii)}{\leq} \Lip(\cT) \|z^{(t)} - \zfp\| + 2
  \|r^{(t)}\| \nonumber \\
  \label{EqnForInduction}
& \stackrel{\rm (iii)}{\leq} \rho \|z^{(t)} - \zfp\| + 2
  \|r^{(t)}\|,
\end{align}
where inequality (i) applies the triangle inequality to the
relation~\eqref{EqnPerturb} between the perturbed and idealized
operators; step (ii) follows by definition of the residual $r^{(t)}$
at round $t$; and step (iii) follows from the
bound~\eqref{ineq:lip-factor} on the Lipschitz coefficient of $\cT$.
Performing induction on this bound yields the stated claim.

\paragraph{Proof of the bound~\eqref{ineq:lip-factor}:}

It remains to bound the Lipschitz coefficient of the idealized
operator $\cT$.  Since the composite function $F(z) \defn \sum_{j=1}^\numdevices f_j(z_j)$
is $\ellmin$-strongly convex and $\Lmax$-smooth, known results on
reflected proximal operators \cite[Theorems 1 and 2]{GisBoy17} imply
that with the stepsize choice $\step = 1/\sqrt{\ellmin \Lmax}$, the
operator $\refl_{\step F}$ satisfies the bound
\begin{align}
  \label{eqn:contraction-refl}
\|\refl_{\step F}(z) - \refl_{\step F}(z')\|_2 & \leq \Big( 1 -
\frac{2}{\sqrt{\kappa} + 1} \Big) \|z - z'\|_2 \qquad \mbox{for all
  $z, z' \in (\real^\usedim)^\numdevices$.}
  \end{align}
On the other hand, the reflected proximal operator $\refl_{I_\eqset}$
for the indicator function $\refl_{I_\eqset}$ is non-expansive, so
that
\begin{align}
\label{EqnNonExpansive}  
  \|\refl_{I_\eqset}(z) - \refl_{I_\eqset}(z)\|_2 & \leq \|z - z'\|_2
  \qquad \mbox{for all $z, z' \in (\real^\usedim)^\numdevices$.}
\end{align}
Applying the triangle inequality and using the
definition~\eqref{EqnIdealized} of the idealized operator $\cT$, we
find that
\begin{align*}
  \|\cT(z) - \cT(z')\|_2 & \leq  \|
  \refl_{\step F} \big( \refl_{I_\eqset}(z) \big) - \refl_{\step F}
  \big( \refl_{I_\eqset}(z') \big)\|_2 \\
& \stackrel{\rm (iv)}{\leq}  \Big(1 -
  \frac{2}{\sqrt{\kappa} + 1} \Big) \; \|\refl_{I_\eqset}(z) -
  \refl_{I_\eqset}(z')\|_2 \\
& \stackrel{\rm (v)}{\leq} \Big( 1 - \frac{2}{\sqrt{\kappa} + 1}
  \Big) \; \|z - z'\|_2,
\end{align*}
where step (iv) uses the contractivity~\eqref{eqn:contraction-refl} of
the operator $\refl_{\step F}$, and step (v) uses the
non-expansiveness~\eqref{EqnNonExpansive} of the operator
$\refl_{I_\eqset}$.  This completes the proof of the
bound~\eqref{ineq:lip-factor}.

  %%%%%%%%%%%%%%%%%%%%%%%%%%%%%%%%%%%%%%%%%%%%%%%%%%%%%%%%%%%%%%%%%%%%%%%%%%%%%%%%%%%%%%%%%%%%%%%%%%%%%%%%%%%%%%
  
\subsubsection{Proof of Corollary~\ref{CorFedSplitStrong}}
  
By construction, the function $h_j$ is smooth with parameter $M \defn
\step \Lmax + 1$ and strongly convex with parameter $m \defn \step
\ellmin + 1$.  Consequently, if we define the operator $H_j(u) \defn u
- \alpha \nabla h_j(u)$, then by standard results on gradient methods
for smooth-convex functions, the stepsize choice $\alpha = \frac{2}{M
  + m}$ ensures that the operator $H_j$ is contractive with parameter
at least $\rho = 1 - \frac{m}{M}$.  Thus, we have the bound
\begin{align*}
\|u^{(\numepoch+1)} - u^*\|_2 & \leq \rho^\numepoch \|u^{(1)} -
\ustar\|_2,
\end{align*}
where $\ustar = \prox_{\step f_j}(x_j^{(t)})$ is the optimum of the
proximal subproblem.  Unpacking the definitions of $(m, M)$ and
recalling that $\step = 1 /\sqrt{\ellmin \Lmax}$, we have
\begin{align*}
\frac{M}{m} = \frac{ \step \Lmax + 1}{\step \ellmin + 1} \; = \;
\frac{\sqrt{\frac{\Lmax}{\ellmin}} + 1}{\sqrt{\frac{\ellmin}{\Lmax}} +
  1} \leq \; \sqrt{\kappa} + 1,
\end{align*}
and hence $\rho \leq 1 - \frac{1}{\sqrt{\kappa} + 1}$, which
establishes the claim.

%%%%%%%%%%%%%%%%%%%%%%%%%%%%%%%%%%%%%%%%%%%%%%%%%%%%%%%%%%%%%%%%%%%%%%%%%%%%%%%%%%%%%%%%%%%%%%%%%%%

\subsubsection{Proof of Theorem~\ref{thm:reduction}}
\label{proof:reduction}

Recalling the definition~\eqref{eqn:regularized-objective} of the
regularized objective $F_\lambda$, note that it is related to the
unregularized objective $F$ via the relation $F_\lambda(x) = F(x) +
\frac{\numdevices \lambda}{2} \|x- x^{(1)}\|^2$, where $x^{(1)}$ is
the given initialization. The proposed procedure is to compute an
approximation to the quantity
\begin{align*}
  x^\star_\lambda \defn \argmin_{x \in \R^d} \Bigg(
  \underbrace{\sum_{j=1}^\numdevices \Big\{ f_j(x) +
    \frac{\lambda}{2}\|x - x^{(1)}\|^2 \Big\}}_{ = : F_\lambda(x)}
  \Bigg).
\end{align*}

Now suppose that we have computed a vector $\xhat \in \real^d$
satisfies $F_\lambda(\xhat) - F_\lambda(x^\star_\lambda) \leq \eps/2$.
Letting $\Fstar = F(\xstar)$ denote the optimal value of the original
(unregularized) optimization problem, we have
\begin{align}
  \label{EqnStart}
F(\xhat) - \Fstar & = \Big \{ F(\xhat) - F_\lambda(\xstar_\lambda) \Big \} +
\Big \{ F_\lambda(\xstar_\lambda) - F(\xstar) \Big \}.
\end{align}
By definition of $F_\lambda$, we have $F(\xhat) \leq
F_\lambda(\xhat)$.  Moreover, again using the definition of
$F_\lambda$, we have
\begin{align*}
F_\lambda(\xstar_\lambda) - F(\xstar) & = F_\lambda(\xstar_\lambda) -
F_\lambda(\xstar) + \frac{\numdevices \lambda}{2}\|\xstar -
x^{(1)}\|^2 \\
& \leq \frac{\numdevices \lambda}{2}\|\xstar -
x^{(1)}\|^2,
\end{align*}
where the inequality follows since $\xstar_\lambda$ minimizes
$F_\lambda$ by definition.  Substituting these bounds into the initial
decomposition~\eqref{EqnStart}, we find that
\begin{align}
F(\xhat) - \Fstar & \leq \Big \{ F_\lambda(\xhat) -
F_\lambda(\xstar_\lambda) \Big \} + \frac{\numdevices
  \lambda}{2}\|\xstar - x^{(1)}\|^2 \nonumber \\
\label{ineq:eps-suboptimality}
& \leq \frac{\eps}{2} + \frac{\eps}{2} = \eps.
\end{align}
where the inequality follows since since $\xhat$ is
$(\eps/2)$-cost-suboptimal for $F_\lambda$, and by our selection of
$\lambda$.  Thus to finish the proof, we simply need to check how many
iterations it takes to compute an $(\eps/2)$-cost-suboptimal point for
$F_\lambda$.

Let us define the shorthand notation $\Lsum \defn
\sum_{j=1}^\numdevices L_j$ and $\kappa_\lambda \defn \frac{\Lmax +
  \lambda}{\lambda}$ Since $F_\lambda$ is a sum of functions that are
$\lambda$-strongly convex and $(L_j + \lambda)$-smooth, it follows
that from initialization $x^{(1)}$, the \fedsplitspace algorithm
outputs iterates $x^{(t)}$ satisfying the bound
\begin{align}
F_\lambda(x^{(t + 1)}) - F_\lambda(\xstar_\lambda)
& \stackrel{\rm (i)}{\leq}
\frac{\Lsum + m\lambda}{2} \|x^{(t+1)} - \xstar_\lambda\|^2
\nonumber \\
\label{ineq:reg-subopt}
& \stackrel{\rm (ii)}{\leq} \frac{\Lsum + m\lambda}{2} \left(1 -
\frac{2}{\sqrt{\kappa_\lambda} + 1} \right)^{\! 2t} \frac{\|x^{(1)} -
  \zstar_\lambda\|^2}{\numdevices}.
\end{align}
In the above reasoning, inequality (i) is a consequence of the
smoothness of the losses $f_j$ when regularized by $\lambda$, along
with the first-order optimality condition for $\xstar_\lambda$; and
bound (ii) then follows by squaring the guarantee of
Theorem~\ref{thm:inexact-convergence} with $b = 0$.  By inverting the
bound~\eqref{ineq:reg-subopt}, we see that in order to achieve an
$\eps/2$-optimal solution, it suffices to take the number of
iterations $t$ to be lower bounded as
\begin{align*}
t \geq \ceil{\frac{\sqrt{\kappa_\lambda} + 1}{4} \log\left\{
  \frac{(\Lsum + \lambda m) \|x^{(1)} -
    \zstar_\lambda\|^2}{\numdevices} \right\} }.
\end{align*}
Evaluating this bound with the choice $\kappa_\lambda = 1 +
\Lmax/\lambda$ and recalling the bound~\eqref{ineq:eps-suboptimality}
yields the claim of the theorem.

%%%%%%%%%%%%%%%%%%%%%%%%%%%%%%%%%%%%%%%%%%%%%%%%%%%%%%%%%%%%%%%%%%%%%%%%%%%%%%%%%%%%%%%%%%%%%

\subsection{Characterization of fixed points}

In this section we give the two fixed point results for
\texttt{FedSGD} and \texttt{FedProx} as stated in
Section~\ref{sec:framework}.

\subsubsection{Proof of Proposition~\ref{prop:FedGD-fixed-points}}
\label{proof:FedGD-fixed-points}

We begin by characterizing the fixed points of the \texttt{FedSGD}
algorithm.  By definition, any limit point $(x_1^\star, \dots,
x_\numdevices^\star) \in (\real^d)^\numdevices$ must satisfy the fixed
point relation
\begin{align*}
\xstar_j = \frac{1}{\numdevices}\sum_{j=1}^\numdevices
\gradj^{\numepochs}(\xstar_j), \qquad j = 1, 2, \ldots, \numdevices.
\end{align*}
Thus, the limits $x_j^\star$ are common, and this gives part (a) of
the claim.  Expanding the iterated operator $\gradj^{\numepochs}$
gives part (b).

%%%%%%%%%%%%%%%%%%%%%%%%%%%%%%%%%%%%%%%%%%%%%%%%%%%%%%%%%%%%%%%%%%%%%%%%%%%%%%%%%%%%%%%%%%%%%%%%

\subsubsection{Proof of Proposition~\ref{prop:FedProx-fixed-points}}
\label{proof:FedProx-fixed-points}

We now characterize the fixed points of the \texttt{FedProx}
algorithm.  By definition, any limit point $(x_1^\star, \dots,
x_\numdevices^\star)$ satisfies
\begin{align}
  \label{eqn:fedprox-expansion}
\xstar_j = \frac{1}{\numdevices} \sum_{j=1}^\numdevices \prox_{\step
  f_j}(\xstar_j), \qquad j =1, 2, \ldots, \numdevices.
\end{align}
Thus, the limits $x_j^\star$ are common, and this gives part (a) of
the claim.

For any convex function, $f \colon \R^d \to \R$, the proximal operator
satisfies
\begin{align*}
\prox_{\step f}(v) = v - \step \nabla M_{\step f}(v), \quad \text{for
  all}~\step > 0~\text{and}~v \in \R^d.
\end{align*}
Using this identity in display~\eqref{eqn:fedprox-expansion} yields
part (b) of the claim.

%%%%%%%%%%%%%%%%%%%%%%%%%%%%%%%%%%%%%%%%%%%%%%%%%%%%%%%%%%%%%%%%%%%%%%%%%%%%%%%%%%%%%%%%%%%%%%%%%

\section{Discussion}
\label{sec:discussion}

In this paper, we have studied the problem of federated optimization,
in which the goal is to minimize a sum of functions, with each
function assigned to a client, using a combination of local updates at
the client and a limited number of communication rounds.  We began by
showing that some previously proposed methods for federated
optimization, even when considered in the simpler setting of convex
and deterministic updates, need not have fixed points that correspond
to the optima of original problem.  We addressed this issue by
proposing and analyzing a new scheme known as \fedsplit, based on
operator splitting.  We showed that it that does indeed retain the
correct fixed points and we provided convergence guarantees for
various forms of convex minimization problems
(Theorems~\ref{thm:inexact-convergence} and~\ref{thm:reduction}).

This paper leaves open a number of questions.  First of all, the
analysis of this paper was limited to deterministic algorithms,
whereas in practice, one typically uses stochastic approximations to
gradient updates.  In the context of the \fedsplitspace procedure, it
is natural to consider stochastic approximations to the proximal
updates that underlie it.  Given our results on the incorrectness of
previously proposed methods and the work of Woodworth and
colleagues~\cite{WooEtAl20} on the suboptimality on multi-step
stochastic gradient methods, it is interesting to develop a precise
characterization of the tradeoff between the accuracy of stochastic
and deterministic approximations to intermediate quantities and rates
of convergence in federated optimization.  It is also interesting to
consider stochastic approximation methods that exploit higher-order
information, such as the Newton sketch algorithm and other second-order
subsampling procedures~\cite{PilWai16a,PilWai17_NS}.

Moreover, the current paper assumed that updates at all clients are
performed synchronously, whereas in federated problems, often the
client updates are carried out asynchronously.  We thus intend to
explore how the
\fedsplitspace and related splitting procedures behave under
stragglers and delays in computation.  Finally, an important
desideratum in federated learning is that local updates are carried
out with suitable privacy guarantees for the local
data~\cite{BhoEtAl18}.  Understanding how noise aggregated through
differentially private mechanisms couples with our inexact convergence
guarantees is  a key direction for future work.

\subsection*{Acknowledgements}

We thank Bora Nikolic for his careful reading and comments on an
initial draft of this manuscript.
RP was partially supported by a Berkeley ARCS Fellowship.  MJW was
partially supported by Office of Naval Research grant
DOD-ONR-N00014-18-1-2640, and NSF grant NSF-DMS-1612948.

%%%%%%%%%%%%%%%%%%%%%%%%%%%%%%%%%%%%%%%%%%%%%%%%%%%%%%%%%%%%%%%%%%%%%%%%%%%%%%%%%%%%%%%%%%%%%%%%%%%%%%%%

\bibliographystyle{plain}
\bibliography{references}
%%%%%%%%%%%%%%%%%%%%%%%%%%%%%%%%%
\end{document}